%% file: main.tex
\documentclass[journal]{IEEEtran} 

\input{includes}

\begin{document}

\author{Niklas Hargus$^{1}$ and Andreas Orthey$^{1}$ and Marc Toussaint$^{1,2}$}
\title{\Huge Coordinated Multi-Robot Disassembly for Makespan Optimization of Large-Scale Assemblies}
\input{src/figures/figure_page1_pullfigure.tex}
{
  \footnotetext[1]{Technical University of Berlin, Germany}
  \footnotetext[2]{Robotics Institute Germany (RIG)}
}

\input{src/00_abstract}
\input{src/01_introduction}
\input{src/02_related_work}

\input{src/03_method}
\input{src/04_tasks}
\input{src/05_collisionchecking}
\input{src/06_experiments}
\input{src/07_conclusion}

\bibliographystyle{IEEEtranSN}

{\footnotesize
\balance
\bibliography{bib/general.bib}
}

\end{document}

%% file: includes.tex
\usepackage{graphicx}
\usepackage{url}
\usepackage{amsmath}
\usepackage{siunitx}
\usepackage{amssymb}
\usepackage{hyperref}
\usepackage{float}
\usepackage{forest}
\usepackage{balance}\usepackage{amsmath, amssymb, amsthm}
\usepackage[numbers]{natbib}

\usepackage[linesnumbered,ruled,vlined]{algorithm2e}
\SetAlgoNlRelativeSize{-1}   
\SetNlSty{}{}{.}
\SetKwComment{Comment}{// }{}
\SetKwInput{KwData}{Input}     
\SetKwInput{KwParam}{Parameters}
\SetKwInput{KwResult}{Output}   
\DontPrintSemicolon
\usepackage{xcolor}

\usepackage{subcaption}

\usepackage{tcolorbox}
\newcommand{\sqbox}[1]{{\textcolor{#1}{\rule{1.5ex}{1.5ex}}}}

\usepackage{tikz}
\usetikzlibrary{patterns, shapes.geometric, arrows.meta, positioning, calc, fit, backgrounds, shadows}

\usepackage[font=small, labelfont=bf]{caption}

%% file: src/figures/figure_page1_pullfigure.tex
\twocolumn[{%
\begin{@twocolumnfalse}
\centering
\maketitle
\includegraphics[width=0.33\textwidth, height=0.28\textwidth]{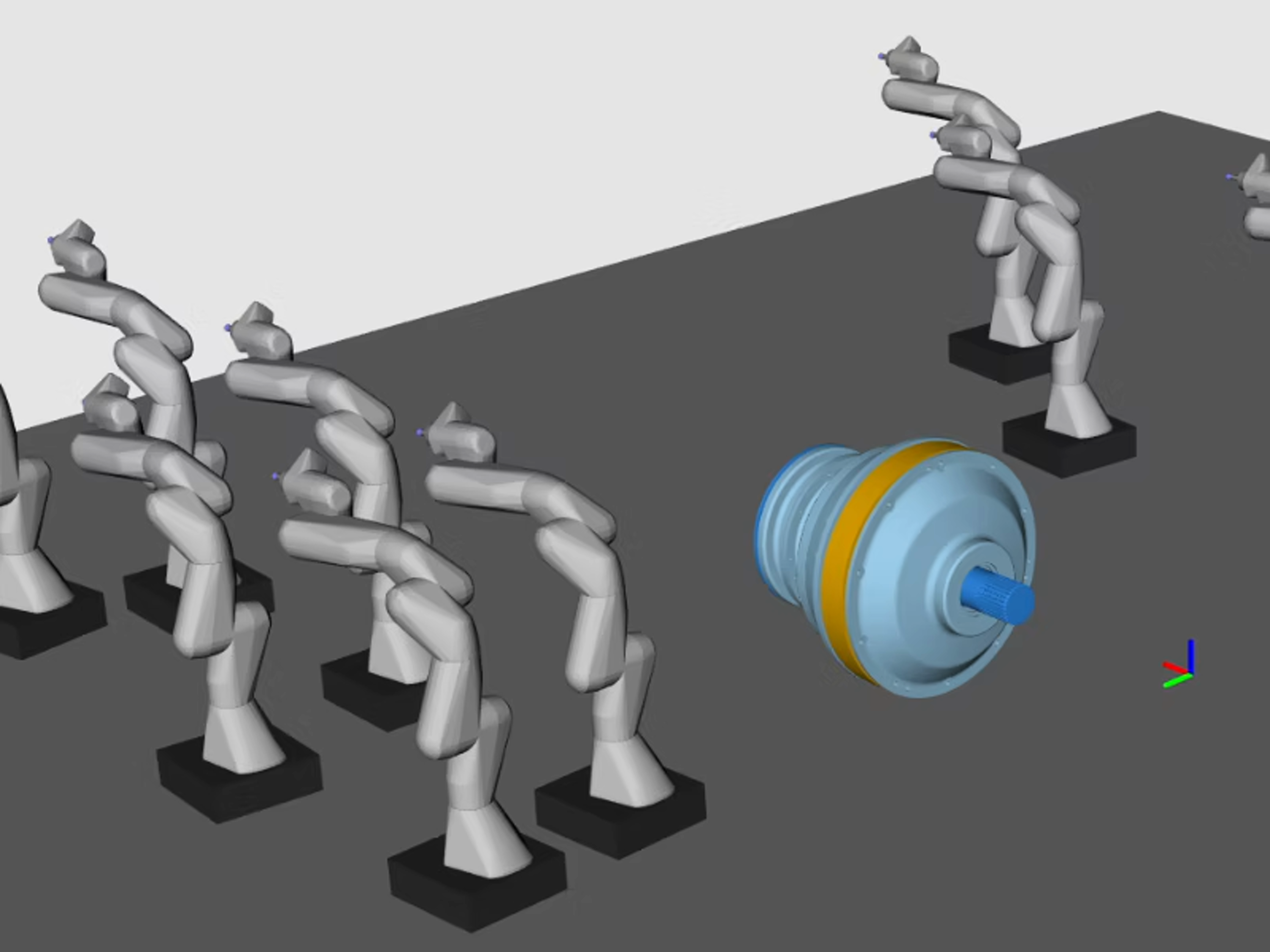}
\includegraphics[width=0.33\textwidth, height=0.28\textwidth]{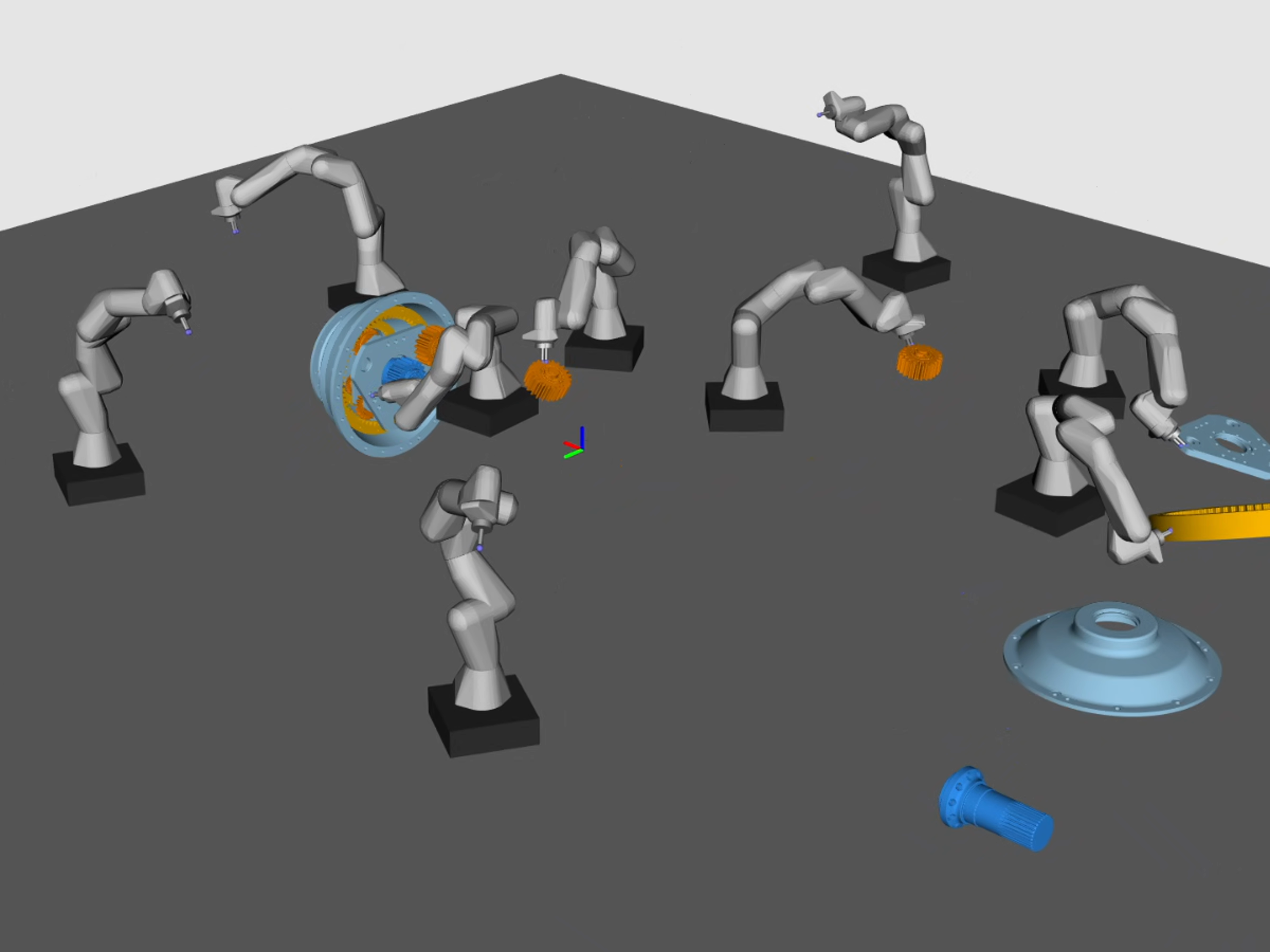}
\includegraphics[width=0.32\textwidth, height=0.28\textwidth]{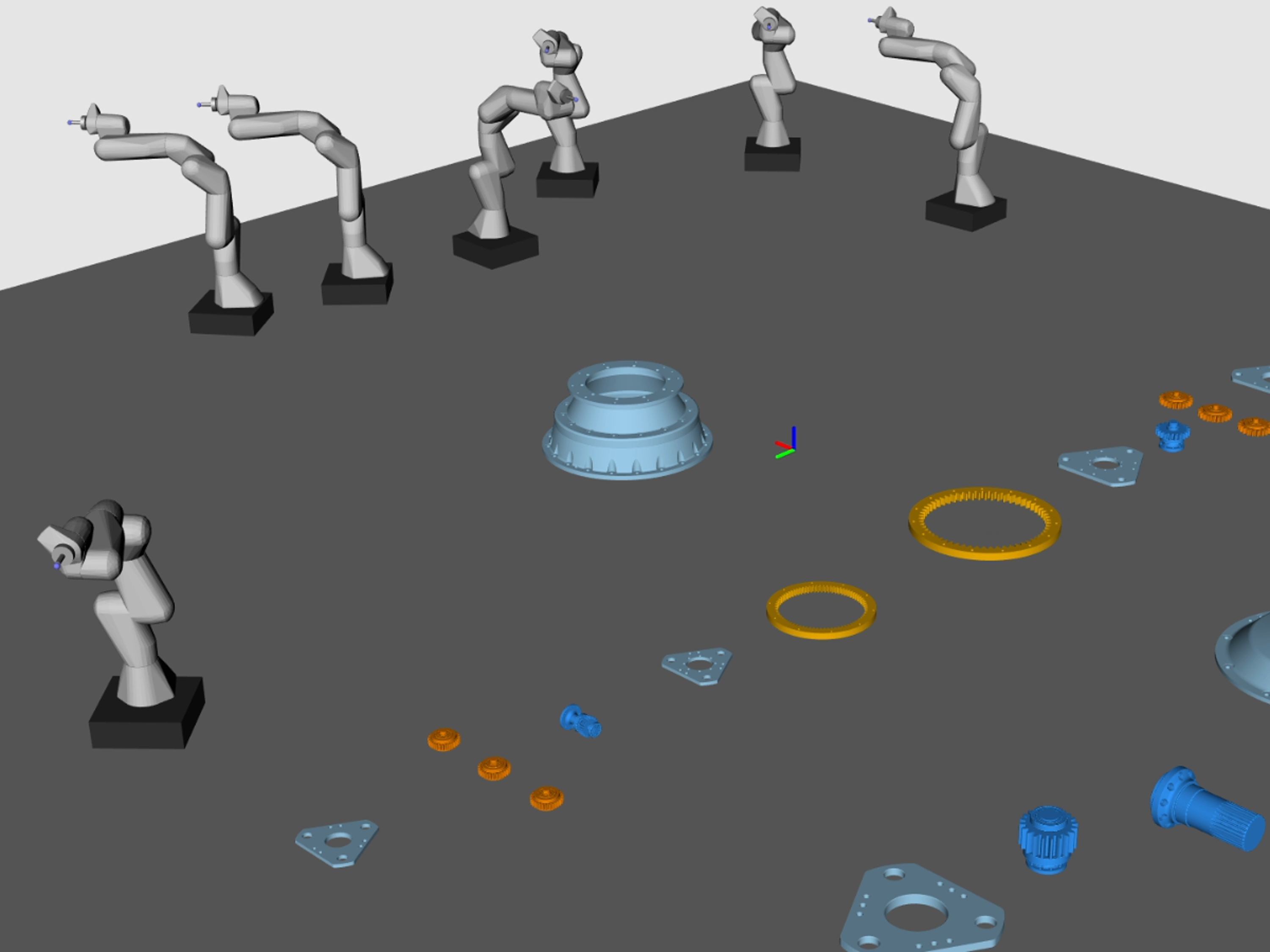}
\captionof{figure}{We develop a coordinated disassembly multi-robot planner to solve large-scale multi-robot task
  and motion planning problems. Pictured is our framework
  coordinating $9$ mobile manipulators which are disassembling a gearbox with
  $33$ pieces. \textbf{Left:} Gearbox in assembled state, robots are approaching
  pick positions. \textbf{Middle:} Robots during disassembly with some robots
  transporting parts, while others are returning for the next pick.
  \textbf{Right:} Final disassembled state of gearbox with robots idle and
  object parts spread out on the floor.\label{fig:pullfigure}}
\end{@twocolumnfalse}
}]

%% file: src/00_abstract.tex
\begin{abstract}
    Multi-robot task and motion planning for disassembly tasks requires robots
    to operate in confined workspaces while coordinating their motions with
    other robots. To tackle this problem, we propose a planning method called coordinated multi-robot disassembly (CoMuDi). CoMuDi coordinates a team of robots for disassembly tasks. The input is a team of robots, an assembly of objects, and a dependency graph. Based on this information, we create compound tasks for pick, place, and exit motions. By propagating temporal constraints, we ensure that each robot can start and end their tasks as early as possible while avoiding collisions with nearby robots. By integrating the space-time RRT* planner (ST-RRT*) into CoMuDi, we ensure that individual tasks minimize arrival time and thereby help us minimize overall makespan. We compare the performance of CoMuDi using both ST-RRT* and RRT* planners with varying time bounds, demonstrating that the combination of CoMuDi and ST-RRT* leads to a higher success rate while minimizing makespan. Finally, we evaluate CoMuDi on six assemblies with up to 49 pieces and up to 9 robots. In those scenarios, we show that CoMuDi returns robot paths that exhibit low idle times, thereby demonstrating that CoMuDi can reliably solve large-scale assemblies. Videos can be found on \href{https://sites.google.com/view/CoMuDi}{https://sites.google.com/view/CoMuDi}.
\end{abstract}

%% file: src/01_introduction.tex
\section{Introduction}
\input{src/figures/figure_system}

Coordinating multiple robots to disassemble objects is a fundamental skill for automating industries like recycling, repair, and remanufacturing. 
However, coordinating the motion and task assignments of robot teams is difficult. 
This is caused by robots needing to work in close proximity (e.g., multiple arms removing pieces from the same battery). 
Furthermore, the robots have to coordinate their tasks in time so that they minimize their idle time and decrease the total execution time. 

To tackle those problems, we develop the coordinated multi-robot disassembly planner (CoMuDi). This algorithm combines existing ideas from multi-robot assembly~\cite{LongHorizon} to coordinate task assignment, from scale-invariant sampling~\cite{Bayraktar2026WAFR} to remove objects from tight, narrow passages, and from space-time planning~\cite{STRRTstar} to minimize arrival time in individual tasks. This is supplemented by our main contributions, which involve a prioritized task queue to ensure that tasks are executed in the correct order and a temporal constraint propagation scheme to allow for more efficient coordination in time. The outcome of this planner are collision-free trajectories to disassemble arbitrary input objects, as is visualized in Fig.~\ref{fig:pullfigure}.

To summarize, our contributions include:
\begin{itemize}
    \item We develop the coordinated multi-robot disassembly planner (CoMuDi), a prioritized sequential task and motion planning framework for multi-robot disassembly to minimize makespan. The framework features task queues to schedule and balance workloads, and a temporal constraint propagation mechanism to minimize idle time.
    \item We introduce a generalized disassembly task, which combines robot navigation, attach and detach tasks, object extraction and transport, and exit tasks into one cohesive task over which we minimize arrival time.
    \item We develop a collision checking framework that tracks attachment data and checks for future collisions to ensure that every robot and object is collision free.
    \item We implement the proposed framework in OMPL using ST-RRT*~\cite{STRRTstar} as the underlying motion planner to handle unknown arrival times.
\end{itemize}

%% file: src/figures/figure_system.tex
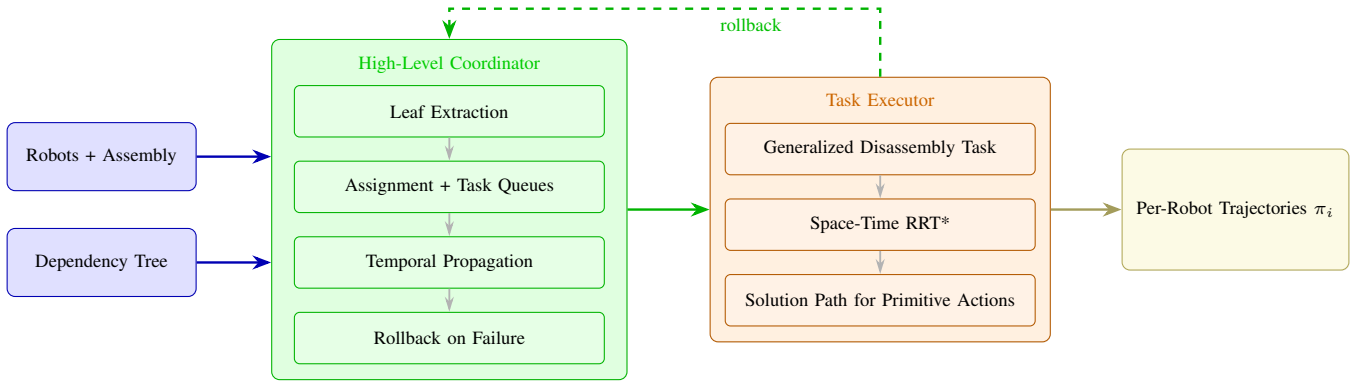
\begin{figure*}[t]
    \centering
\begin{tikzpicture}[
  node distance=0.6cm and 1.0cm,
  box/.style={
    rectangle, draw=#1!70!black, fill=#1!12,
    rounded corners=3pt, align=center,
    minimum width=2.7cm, minimum height=0.9cm,
    font=\scriptsize, inner sep=4pt,
    drop shadow={shadow xshift=0.3pt, shadow yshift=-0.3pt, opacity=0.1}
  },
  smallbox/.style={
    rectangle, draw=#1!70!black, fill=#1!10,
    rounded corners=2pt, align=center,
    minimum width=4.1cm, minimum height=0.68cm,
    font=\scriptsize, inner sep=3pt
  },
  arrow/.style={-{Stealth[length=2mm, width=1.2mm]}, line width=0.7pt, >=stealth},
  thickarrow/.style={-{Stealth[length=2.5mm]}, line width=0.9pt, >=stealth, draw=#1!70!black},
  label/.style={font=\tiny, align=center, text=#1!75!black},
]
\node[box=blue, minimum width=2.5cm] (robots) at (1.2, 4.2) {Robots + Assembly};
\node[box=blue, minimum width=2.5cm] (depgraph) at (1.2, 2.8) {Dependency Tree};

\node[box=green, minimum width=4.7cm, minimum height=4.5cm] (coordinator) at (5.8, 3.5) {};
\node[font=\scriptsize, anchor=north, text=green!80!black] at ([yshift=-0.1cm]coordinator.north) {High-Level Coordinator};

\node[smallbox=green] (leafs) at (5.8, 4.8) {Leaf Extraction};
\node[smallbox=green] (assign) at (5.8, 3.8) {Assignment + Task Queues};
\node[smallbox=green] (temporal) at (5.8, 2.8) {Temporal Propagation};
\node[smallbox=green] (rollback) at (5.8, 1.8) {Rollback on Failure};

\node[box=orange, minimum width=4.5cm, minimum height=3.5cm] (executor) at (11.5, 3.5) {};
\node[font=\scriptsize, anchor=north, text=orange!80!black] at ([yshift=-0.1cm]executor.north) {Task Executor};

\node[smallbox=orange] (primitives) at (11.5, 4.3) {Generalized Disassembly Task};
\node[smallbox=orange] (keyframes) at (11.5, 3.3) {Space-Time RRT*};
\node[smallbox=orange] (planner) at (11.5, 2.3) {Solution Path for Primitive Actions};

\node[box=yellow!90!black, minimum width=3.0cm, minimum height=1.6cm] (output) at (16.2, 3.5) {Per-Robot Trajectories $\pi_i$};

\draw[thickarrow=blue] (depgraph.east) -- (coordinator.west |- depgraph.east);
\draw[thickarrow=blue] (robots.east) -- (coordinator.west |- robots.east);

\draw[arrow, gray!60] (leafs.south) -- (assign.north);
\draw[arrow, gray!60] (assign.south) -- (temporal.north);
\draw[arrow, gray!60] (temporal.south) -- (rollback.north);

\draw[thickarrow=green] (coordinator.east) -- (executor.west);

\draw[arrow, gray!60] (primitives.south) -- (keyframes.north);
\draw[arrow, gray!60] (keyframes.south) -- (planner.north);

\draw[thickarrow=yellow!80!black] (executor.east) -- (output.west);

\draw[thickarrow=green, dashed] (executor.north) -- ++(0,0.9) -|
node[label=green, pos=0.15, below, font=\scriptsize] {rollback} (coordinator.north);

\end{tikzpicture}
    \caption{Overview of the temporal multi-robot TAMP framework for disassembly
    tasks. As input, we require a team of robots and the assembly consisting of
    objects, together with a dependency tree specifying the possible execution
    order of the objects. This tree is used in the high-level coordinator to
    extract the next possible tasks by extracting the leaves. 
    \label{fig:systems-overview}}
\end{figure*}

%% file: src/02_related_work.tex
\section{Related Work\label{sec:related-work}}

Our framework for multi-robot task and motion planning for disassembly problems
builds upon multiple related approaches. This involves time-optimal motion
planning with unknown arrival times, multi-robot navigation, task and motion
planning, and disassembly sequence planning. We review those topics and discuss
their relation to our work.

\subsection{Time-Optimal Motion Planning}




At the core of our approach lies the problem of time-optimal planning in a combined space-time space~\cite{hsu2002randomized}. A particularly important problem hereby is to avoid dynamic obstacles with known obstacle trajectories.
Approaches to tackle this problem involve roadmap-based dynamic planning approaches for arrival-time minimization~\cite{vandenberg2005roadmap, vandenberg2008timeoptimal} or safe
interval path planning~\cite{phillips2011sipp}, which represents the time dimension through intervals during
which a configuration remains collision-free.
The single-query tree alternative
for space-time planning was first introduced by Sintov and Shapiro \cite{sintov2014timebased}
with Time-Based RRT (TB-RRT) for rendezvous planning of two dynamic systems.
However, it is limited by a fixed arrival time and lacks any optimality guarantees.
Grothe et al.'s~\cite{STRRTstar} extension to RRT*~\cite{karaman2011sampling} treats time itself
as a planning dimension. This allows RRTs to directly plan in space-time
to find asymptotically optimal solutions with respect to arrival time.
It grows a bidirectional tree from the start and finish at an unknown arrival time
with a progressively widening time horizon.
Our work is complementary to time-optimal motion planning as we rely on ST-RRT*~\cite{STRRTstar} as an individual planner to optimize generalized disassembly tasks.

\subsection{Multi-Robot Motion Planning}
The two major approaches to multi-robot motion planning are separated by how the
robots' individual configuration spaces are coupled. The centralized approach
combines all configuration spaces together, which increases the complexity
of the problem but remains optimal and complete, whereas the decoupled
methods remain scalable, sacrificing completeness and optimality.

The discrete problem
of Multi-Agent Path Finding (MAPF), stated by Yu and LaValle \cite{yu2013intractability},
for time- and distance-optimal MAPF on connected, undirected graphs, is
NP-hard. The difficulties of the naive planning problem motivated Sharon et al. \cite{Sharon2015CBS}
to introduce Conflict-Based Search (CBS), a two stage decomposition into a high-level
conflict-tree and low-level search in which conflicts are solved. CBS is optimal and complete
on a grid-world and has grounded a large family of multi-agent path finders.
The graph search approach of Phillips et al. \cite{phillips2011sipp} was extended
by Andreychuk et al. \cite{andreychuk2022multi} to multi-agent pathfinding with
continuous time. This approach struggles to extend well to the high-dimensional
configuration spaces of articulated manipulators.

Sampling-based multi-robot planning works in continuous, high-dimensional configuration spaces.
Wagner and Choset \cite{wagner2011mstar} introduced M*, which searches
an implicit representation of the combined roadmaps rather than constructing the joined
roadmap, resulting in significant speedup. However, it is inefficient for higher
dimensions due to the exponential growth of neighbors. This led to
Solovey et al. \cite{dRRTjournal} introducing discrete-RRT (dRRT), which extends
M* with an oracle function to pick the best neighbor.
Shome et al. \cite{dRRTstar} contributed the
asymptotically optimal extension dRRT*, resulting in convergence to an optimum
even in multi-robot manipulator spaces.

Erdmann and Lozano-Pérez~\cite{erdmann} introduced prioritized planning to
circumvent the complexity increase. They propose to
assign an ordering to the robots and plan each one in turns against higher-priority
robots, who are treated as dynamic obstacles in time. This approach trades
completeness and optimality for gain in tractability.

Bennewitz et al. \cite{bennewitz2002finding} demonstrated that fixed orderings
can fail to find valid solutions, even though they exist. Čáp et al. \cite{cap2015prioritized}
provides a characterization instance for which a revised prioritized algorithm
is complete. Priority-Based Search (PBS) by Ma et al. \cite{ma2019pbs} bridges prioritized planning
and CBS by searching over partial orderings.
However, these claims are only evaluated for mobile robots, rather than manipulators.

Our approach differs in that we focus on general task and motion planning, not solely on multi-robot navigation.

\subsection{Task and Motion Planning}

Task and Motion Planning (TAMP) addresses the problem in which a robot must
simultaneously decide on what to do at a symbolic level and how to do it at a
geometric level. The survey by Garrett et al. \cite{garrett2021survey} distinguishes
into three broad paradigms: classical hierarchical interface-based, sampling-based
multi-modal, and optimization-based methods.

The classical TAMP formulation uses a symbolic planner to generate candidates.
Kaelbling et al. \cite{Kaelbling2011TAMP} introduced an early work
which provides a general strategy for hierarchical planning and execution by limiting
the search space. Srivastava et al. \cite{scrivastava2014Combined} improves upon existing
task and motion planning solutions by introducing a predicate-level interface
between geometric and symbolic layer. Garrett et al. introduce PDDLStream
\cite{garrett2020pddlstream}, an extension to the
Planning Domain Definition Language (PDDL) \cite{PDDL}, which supports black-box implementations
for grasps, placements and trajectories by providing adaptive algorithms that interleave
symbolic search and constraint satisfaction.

Sampling-based TAMP uses mode switching in hybrid configuration spaces of
intersecting sub-manifolds. Each mode switch amounts to sampling mode transitions
and concatenating intra-mode paths. Siméon et al. \cite{simeon2004manipulation}
establish this perspective with their PRM-based manipulation planner, Hauser
and Latombe \cite{hauser2010multimodal} provide a more modern formalization with
probabilistic completeness. Krontiris and Bekris \cite{krontiris2015rearrangement}
address non-monotone object rearrangement and Vega-Brown and Roy \cite{vegabrown2016asymptotically}
provide the first asymptotically optimal guarantee with piecewise-analytic
differential constraints for TAMP.

Toussaint \cite{lgp} introduced Logic-Geometric Programming (LGP),
in which a symbolic search over action sequences is interleaved with a non-linear
program. This collapses the multi-level approach from previous approaches into
a single optimization problem. KOMO \cite{toussaint2017komo} is used as the
trajectory-level solver, a k-order Markov constrained NLP solved by Gauss-Newton
with augmented Lagrangian. This integrated optimization-based approach was
demonstrated to extend to contact dynamics and tool usage \cite{toussaint2018differentiable}.

\subsection{Multi-Robot Task and Motion Planning}

Multi-robot task and motion planning (MR-TAMP) shares the same structure as
single-robot TAMP with high-level decision making and low-level execution, but
expands to task allocation, motion coordination and often makespan reduction.

Pan et al. \cite{pan2021mrTAMPframework} extended classic PDDL-based TAMP
to multi-robot systems. They introduce a scheduler between the symbolic task
planner and multi-robot motion planner. This scheduler is responsible for reducing
the number of task planner calls and biasing the search based on geometric insights.

Toussaint and Lopes \cite{toussaint2017MultiBound}
generalized LGP to cooperative manipulation between a humanoid and Baxter robot
but do not push to a larger team of robots. Hartmann et al. \cite{LongHorizon}
demonstrates long-horizon multi-robot planning for construction work by
decomposing large scale TAMP problems into smaller one robot subproblems. They
leverage Logic-Geometric Programming \cite{lgp} to solve for keyframes, which are then connected
by ST-RRT* \cite{STRRTstar}. This was followed by formalizing the multi-arm multi-task
planning problem based on greedy descent with random restarts to optimize the
total makespan \cite{hartmann2023towards}.

Chen et al. \cite{chen2022CoopTAMP}
approach multi-arm assembly problems by formulating a high-level mixed-integer program to
solve for the task assignment under precedence constraints and a CBS-based low level
planner to route the robots collision-free, yielding high-quality makespans at
substantial computational cost.

Our work is complementary to those approaches in that we apply MR-TAMP to disassembly problems. While we also use ST-RRT*~\cite{STRRTstar}, similar to Hartmann et al.~\cite{LongHorizon}, our method includes scale-invariant sampling for integrating object extraction tasks. Additionally, we extend this to disassembly tasks by proposing a generalized disassembly task, the propagation of temporal constraints, and an efficient time-dependent collision-checking framework.

\subsection{Disassembly Sequence Planning}
Disassembly through robots connects the field of disassembly sequence planning (DSP) to
robotics, by using the output of DSP as the input for robot task and motion planning.
Lambert's survey \cite{lambert2003DisassemblySequencing} provides the classical
foundation for DSP, which is formalized by graph- and AND/OR-based representations.
Smith et al. \cite{smith2012dssg} extend this representational line with
Disassembly Sequence Structure Graphs (DSSG), which handle multi-target selective disassembly
by encoding precedence and accessibility in a single structure.

The graph-based approach is contrasted by treating DSP as optimization-based problem.
This line of work focuses on genetic algorithms (GA). Kongar et al. \cite{kongar2006ga} and
Lazzerini et al. \cite{lazzerini2000ga} provide early work on GA-DSP, while
Kheder et al. \cite{kheder2014SequenceGeneticAlgo} provide extensions to address
tool changes and accessibility violations under precedence constraints.

The precomputed nature of all DSP approaches sits in contrast to recent
work by Poschmann et al. \cite{poschmann2021fostering}, which
shifts to information-driven and learning-based approaches to acquire
information at runtime about the assembly state.

%% file: src/03_method.tex
\section{Coordinated Multi-Robot Disassembly}

\subsection{Problem Statement\label{sec:problem-statement}}

We consider a workspace $\mathcal{W}$ with $N$ robots and $K$ rigid objects.
Each robot $r_i \in \mathcal{R}=\{r_1,\dots,r_N\}$ has a configuration space 
$\mathcal{Q}_i$ with $d_i$ degrees of freedom.
The compound state space is defined as 
$X = \mathcal{Q}_1 \times \cdots \times \mathcal{Q}_N$ with
a combined dimension $d = \sum d_i$. 
Additionally, there exist $K$ rigid objects 
$\mathcal{O} = \{o_1, \dots, o_K\}$ in $\mathcal{W}$, with each object $o_j \in \mathcal{O}$ being described by a
starting pose $T^j_\text{start} \in SE(3)$ in the assembled state and a
goal pose $T^j_\text{goal} \in SE(3)$ in the disassembled state. 

\paragraph{Dependency Graph}

Additionally, objects are not free to move in any sequence, but there exists 
a dependency graph $G$. $G$ is a directed acyclic graph (DAG) that 
defines the order in which the objects have to be moved to accomplish the
disassembly. $G$ consists of vertices $\boldsymbol{V}$, representing the
objects, and edges $\boldsymbol{E}$, representing the dependencies.
As an example, Fig.~\ref{fig:dependency-graph} shows a dependency graph with
four objects, whereby the panel depends on bolt A and B, which in turn depend on the lid.
All leaf nodes in the graph have no dependencies on other nodes.

\paragraph{Assumptions}
\label{sec:assumptions}
The goal of this paper is to coordinate a multi-robot team for
disassembly. To study this in isolation, we make two assumptions to
simplify the problem:

\begin{itemize}
    \item The end-effector of each robot is modeled as an omnidirectional gripper, enabling grasps from any angle on the surface of the object. This abstracts away the problem of
    grasp planning, which differs depending on the gripper types.
    \item Execution of paths is done in a non-physical simulation. This ensures
      that effects such as stick-slip or object resting behaviors do not have to
      be considered. 
\end{itemize}

This formulation allows us to concentrate on the task and motion planning problem itself. We believe those are reasonable assumptions that allow us to find solutions that act as necessary conditions for solving disassembly problems in the real world~\cite{assembleThemAll}.

\paragraph{Goal}
Given robot and object assembly, we define the \emph{multi-robot disassembly problem} as the problem of finding a set of trajectories
$\mathcal{T}$, with each trajectory $\pi_i \in \mathcal{T}$ being a 
sequence of pairs consisting of a configuration and an associated time point for robot $r_i$. 
They describe the collision free motion for each robot and objects to reach the disassembled state. 

In this paper, we are also interested in the \emph{optimal multi-robot disassembly problem}, which is the problem of finding a set of trajectories $\mathcal{T}^*$ that minimizes the makespan: the total time required for the disassembly process from start to finish.

\input{src/figures/figure_dependency_graph_example.tex}

\subsection{Overview and Implementation}

\input{src/algorithms/algorithm_highlevel}
To solve the optimal multi-robot disassembly problem, we present the coordinated multi-robot disassembly (CoMuDi) planner.
The pseudocode for CoMuDi is described in Alg.~\ref{alg:multirobot-planning}.
The input to the algorithm is a precomputed dependency graph and all available robots.
While the dependency graph is not empty (Line 1), the unassigned leaf nodes are obtained,
which do not depend on any other vertex (Line 2). Next, all robots that
currently do not have an assigned task are selected (Line 3). 
The algorithm then matches an available robot with a node based on an 
assignment policy (Line 6--7). It then creates a task in the robot's task queue 
(Line 8--9) and marks the node as assigned to avoid reassignment (Line 10), continuing until all
possible pairings of leaf-robots have been exhausted (Line 5).
Afterwards, it iterates through every task queue (Line 11), skipping any empty queues 
since no work needs to be done (Line 12--13). The first task in the queue is selected
(Line 14), and the main run of the selected task is invoked (Line 15).
If the task has failed (Line 16), the associated node is unmarked in the tree (Line 17) and is
ready to be reassigned.
If the task is successful, the constraints on the next
node are propagated (Line 19), and the node is removed from the dependency tree 
(Line 20). 

\subsection{Task Queue\label{sec:task-representation}}

Once tasks are assigned to robots, we need to execute them. We assume that each
task is associated with a robot $r_i$, an object $o_j$, a start time
$t_\text{start}$, and a goal region $X_G \subseteq Q_i$, which is represented by
the goal of object $o_j$ (within some threshold).

Centralized storage maintains the relationship between each robot and a queue of tasks
for each robot.
A robot is considered available if and only if its queue is empty.
Tasks are executed in first-in-first-out order.
The front task is invoked at each simulation step until it signals completion.
The completion of the task triggers the removal, and the next task begins
execution. If the queue is empty, the robot is assumed to be available again and ready for
new task assignments.

\subsection{Temporal Dependency Constraints\label{sec:temporal-dependency-constraints}}

The high-level behavior in Alg.~\ref{alg:multirobot-planning}
removes a dependency vertex
after the associated task has been finished. 
Afterwards, new leaf nodes are available for manipulation. This is a conservative approach, since
a successor object can begin its planning before its predecessor has completed
its full task sequence. Waiting for the full completion of the
predecessor's task wastes time, during which a second robot could already begin
approaching the next object.

To exploit this, each vertex in the dependency graph carries a temporal constraint that
controls when temporal bounds are propagated to its successors. This is shown in Fig.~\ref{fig:time-dependencies-pick-place}. 
Without a temporal propagation, the successors become
available only when the vertex is removed from the tree
(Fig.~\ref{fig:time-dependencies-no-time-dep}).
With a temporal constraint, however, the attach and detach times are propagated
(Fig.~\ref{fig:time-dependencies-with-dep}). When the attach time
$t_\text{attach}$ is known, the successor can
finish their pick phase as soon as the predecessor object is picked by another
robot. Similarly, when the detach time $t_\text{detach}$ is known, the successor
can finish their place phase as soon as the predecessor object was placed. Note
that we do not require those constraints to be present, in which case we either
have to wait for the placement (if a dependency exists between objects) or
no constraints are present (if no dependency exists). Having those lower bounds
on the starting times of the respective phases is advantageous to obtain a tight
scheduling without robots being idle.

\input{src/figures/figure_time_dependencies.tex}

%% file: src/figures/figure_dependency_graph_example.tex
\newcommand{\dependencyScalingwidth}{0.48\linewidth}

\begin{figure}[t]
\centering
\begin{subfigure}[b]{\dependencyScalingwidth}
\centering
\resizebox{\linewidth}{!}{%
\begin{tikzpicture}[
    >=Stealth,
    bolt/.style={draw, thick, minimum width=0.2cm, minimum height=1cm, fill=gray!60},
    label/.style={font=\bfseries\sffamily}
]
\def\dentWidth{0.4}
\def\dentHeight{0.8}

\draw[thick, fill=gray!10]
  (-2,0) -- (2,0)
  -- (2,1.2)
  -- (1 + \dentWidth/2, 1.2)
  -- (1 + \dentWidth/2, 1.2 - \dentHeight)
  -- (1 - \dentWidth/2, 1.2 - \dentHeight)
  -- (1 - \dentWidth/2, 1.2)
  -- (-1 + \dentWidth/2, 1.2)
  -- (-1 + \dentWidth/2, 1.2 - \dentHeight)
  -- (-1 - \dentWidth/2, 1.2 - \dentHeight)
  -- (-1 - \dentWidth/2, 1.2)
  -- (-2,1.2)
  -- cycle;

\node[inner sep=0pt, outer sep=0pt, minimum width=4cm, minimum height=1.2cm] (A) at (0,0.6) {};
\node[bolt, label=above:B] (B) at (-1,2.4) {};
\node[bolt, label=above:C] (C) at (1,2.4) {};

\def\yupper{4.0}
\def\ylower{2.3}
\node (L) at (-2.3,\ylower){};
\node (R) at (+2.3,\ylower){};
\draw[thick, fill=gray!20]
  (-2.2,\ylower) -- (-2.4,\ylower) -- (-2.4,\yupper) -- (2.4,\yupper)
  -- (2.4,\ylower) -- (2.2,\ylower) -- (2.2,\yupper-0.5) -- (-2.2,\yupper-0.5) -- cycle;

\draw[thick, dashed, ->] (B.south) -- ++(0,-1cm);
\draw[thick, dashed, ->] (C.south) -- ++(0,-1cm);
\draw[thick, dashed, ->] (L.south) -- ++(0,-1.5cm);
\draw[thick, dashed, ->] (R.south) -- ++(0,-1.5cm);
\end{tikzpicture}%
}
\caption{Assembly}
\end{subfigure}
\hfill
\begin{subfigure}[b]{\dependencyScalingwidth}
\centering
\resizebox{\linewidth}{!}{%
\begin{tikzpicture}[
  every node/.style={
    draw,
    circle,
    very thick,
    minimum size=1.3cm,
    align=center,
    inner sep=2pt
  },
  >=Stealth
]
\node (Panel) at (0,3) {Panel};

\node[below left=0.5cm and 0.5cm of Panel] (BA) {Bolt A};
\node[below right=0.5cm and 0.5cm of Panel] (BB) {Bolt B};
\node[below left=0.5cm and 0.5cm of BB] (L) {Lid};

\draw[->, very thick] (Panel) -- (BA);
\draw[->, very thick] (Panel) -- (BB);
\draw[->, very thick] (BA) -- (L);
\draw[->, very thick] (BB) -- (L);
\end{tikzpicture}%
}
\caption{Dependency Graph}
\end{subfigure}

\caption{Example of an assembly with a lid, two bolts, and panel. The dependency graph shows the assembly constraints.}
\label{fig:dependency-graph}
\end{figure}
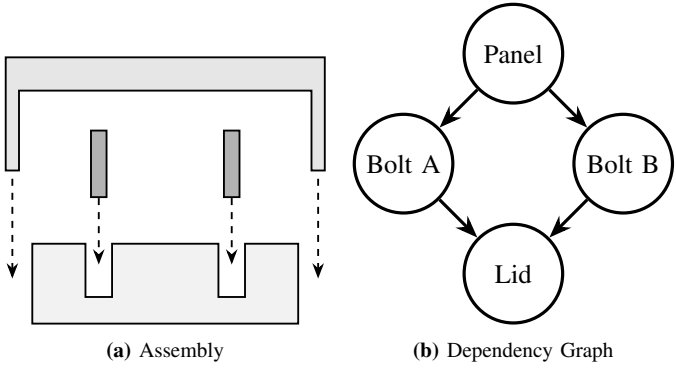

%% file: src/algorithms/algorithm_highlevel.tex
\begin{algorithm}[t]
  \caption{Coordinated Multi-Robot Disassembly}\label{alg:multirobot-planning}
  \KwData{dGraph, robots}
  \KwResult{---}
  
  \While{dGraph $\ne \varnothing$}{
    leaves $\gets$ \texttt{getUnassignedLeaves(dGraph)}\;
    freeRobots $\gets \{ r \in \text{robots} \mid taskQueue(r) = \varnothing \}$\;
    assignedTasks $\gets \varnothing$
    
    \While{leaves $\ne \varnothing \ \wedge\ $ freeRobots $\neq \varnothing$}{
      robot $\gets$ \texttt{pop(freeRobots)}\;
      leaf $\gets$ \texttt{pop(leaves)}\;
      task $\gets$ \texttt{createTask(robot, leaf)}\;
      \texttt{append(assignedTasks, task)}\;
      markLeafAssigned(leaf)\;
    }
    
    \ForEach{$(robot, queue) \in$ assignedTasks}{
      \If{queue $= \varnothing$}{
        \textbf{continue}\;
      }
      currentTask $\gets$ \texttt{pop(queue)}\;
      \texttt{run(currentTask)}\;
      
      \If{currentTask.failed}{
        \texttt{markLeafUnassigned(leaf)}\;
        \textbf{continue}\;
      }
      \texttt{propagateConstraints(dGraph, leaf)}\;
      \texttt{removeVertex(dGraph, leaf)}\;
    }
  }
\end{algorithm}

%% file: src/figures/figure_time_dependencies.tex
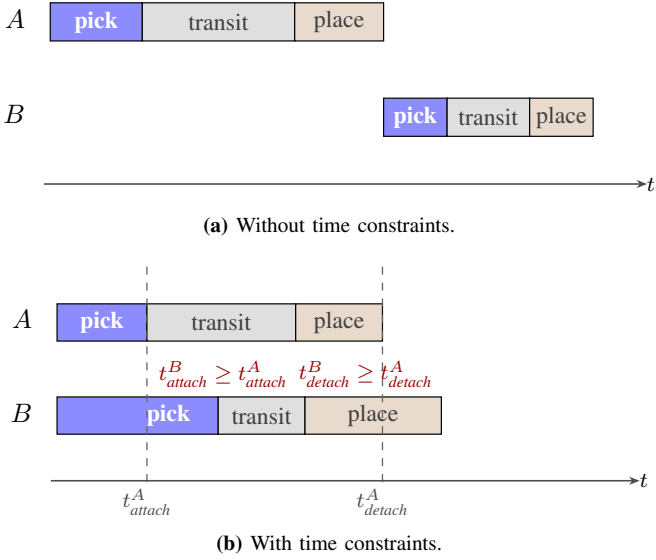
\begin{figure}[t]
\begin{subfigure}[t]{\linewidth}
  \centering
  \begin{tikzpicture}[
    scale=0.84,
    pickph/.style={fill=blue!45,draw=black,line width=0.45pt},
    transitph/.style={fill=gray!25,draw=black,line width=0.45pt},
    placeph/.style={fill=orange!55!black!20,draw=black,line width=0.45pt},
    rowlbl/.style={font=\bfseries,anchor=east},
    phaselbl/.style={font=\small\bfseries,text=white,anchor=center},
    phaselbldark/.style={font=\small,text=black!75,anchor=center},
    eventlbl/.style={font=\footnotesize\itshape,anchor=south,inner sep=1pt,text=black!70},
    taxis/.style={-{Stealth[length=4pt]},black!70,line width=0.6pt},
  ]
  \node[rowlbl] at (-0.25,1.6) {$A$};
  \draw[pickph] (0,1.25) rectangle (1.45,1.85);
  \node[phaselbl] at (0.725,1.55) {pick};
  \draw[transitph] (1.45,1.25) rectangle (3.85,1.85);
  \node[phaselbldark] at (2.65,1.55) {transit};
  \draw[placeph] (3.85,1.25) rectangle (5.25,1.85);
  \node[phaselbldark] at (4.55,1.55) {place};

  \node[rowlbl] at (-0.25,0.1) {$B$};
  \draw[pickph] (5.25,-0.25) rectangle (6.25,0.35);
  \node[phaselbl] at (5.75,0.05) {pick};
  \draw[transitph] (6.25,-0.25) rectangle (7.55,0.35);
  \node[phaselbldark] at (6.9,0.05) {transit};
  \draw[placeph] (7.55,-0.25) rectangle (8.55,0.35);
  \node[phaselbldark] at (8.05,0.05) {place};

  \draw[taxis] (-0.1,-1.0) -- (9.4,-1.0);
  \node[font=\footnotesize\itshape,anchor=west] at (9.25,-1.0) {$t$};
  \end{tikzpicture}
  \caption{Without time constraints.}
  \label{fig:time-dependencies-no-time-dep}

\end{subfigure}
\begin{subfigure}[t]{\linewidth}
  \centering
  \begin{tikzpicture}[
    scale=0.82,
    pickph/.style={fill=blue!45,draw=black,line width=0.45pt},
    transitph/.style={fill=gray!25,draw=black,line width=0.45pt},
    placeph/.style={fill=orange!55!black!20,draw=black,line width=0.45pt},
    rowlbl/.style={font=\bfseries,anchor=east},
    phaselbl/.style={font=\small\bfseries,text=white,anchor=center},
    phaselbldark/.style={font=\small,text=black!75,anchor=center},
    eventlbl/.style={font=\footnotesize\itshape,anchor=south,inner sep=1pt,text=black!70},
    eventline/.style={black!60,dashed,line width=0.5pt},
    taxis/.style={-{Stealth[length=4pt]},black!70,line width=0.6pt},
  ]
  \node[rowlbl] at (-0.25,1.6) {$A$};
  \draw[pickph] (0,1.25) rectangle (1.45,1.85);
  \node[phaselbl] at (0.725,1.55) {pick};
  \draw[transitph] (1.45,1.25) rectangle (3.85,1.85);
  \node[phaselbldark] at (2.65,1.55) {transit};
  \draw[placeph] (3.85,1.25) rectangle (5.25,1.85);
  \node[phaselbldark] at (4.55,1.55) {place};

  \node[rowlbl] at (-0.25,0.1) {$B$};
  \draw[pickph] (0.0,-0.25) rectangle (2.6,0.35);
  \node[phaselbl] at (1.8,0.05) {pick};
  \draw[transitph] (2.6,-0.25) rectangle (4.0,0.35);
  \node[phaselbldark] at (3.3,0.05) {transit};
  \draw[placeph] (4.0,-0.25) rectangle (6.2,0.35);
  \node[phaselbldark] at (5.1,0.05) {place};

  \draw[eventline] (1.45,2.45) -- (1.45,-1.15);
  \node[eventlbl] at (1.45,-1.6) {$t_{\text{attach}}^{A}$};
  \draw[eventline] (5.25,2.45) -- (5.25,-1.15);
  \node[eventlbl] at (5.25,-1.6) {$t_{\text{detach}}^{A}$};

  \node[font=\footnotesize\itshape,text=red!60!black,anchor=west,inner sep=2pt]
    at (1.55,0.72) {$t^B_{\text{attach}} \geq t^A_{\text{attach}}$};
  \node[font=\footnotesize\itshape,text=red!60!black,anchor=east,inner sep=2pt]
    at (6.15,0.72) {$t^B_{\text{detach}} \geq t^A_{\text{detach}}$};

  \draw[taxis] (-0.1,-1.0) -- (9.4,-1.0);
  \node[font=\footnotesize\itshape,anchor=west] at (9.25,-1.0) {$t$};
  \end{tikzpicture}
  \caption{With time constraints.}
\label{fig:time-dependencies-with-dep}

\end{subfigure}
\caption{Example timeline with pick/place time constraints for two robots A and B.}
\label{fig:time-dependencies-pick-place}
\end{figure}

%% file: src/04_tasks.tex
\section{Generalized Disassembly Task}

\input{src/algorithms/task_generalized_disassembly}

While CoMuDi can deal with arbitrary input tasks, we focus in this work on disassembly and define a generalized disassembly task. This task contains a sequence of subtasks as follows:
\begin{enumerate}
  \item \textbf{Move to Object Grasp} Plan a path to a grasp
    configuration on the object surface with unconstrained orientation
  \item \textbf{Attach}: Establish a kinematic coupling between end-effector
    and object.
  \item \textbf{[Optional] Object Removal}: Advance through a set of object poses
    $T_1, \ldots, T_m$ using iterative IK with time scaling.
  \item \textbf{Move to Goal or Insertion Start} Plan a path to transport
    the object to its goal or insertion start pose $T_{\text{goal}}$.
  \item \textbf{[Optional] Object Insertion}: Advance through a set of object poses
    $T'_1, \ldots, T'_k$ to guide the object into its goal position.
  \item \textbf{Detach}: Terminate the kinematic coupling.
  \item \textbf{Exit} Move the robot outside the active workspace to unblock any other robots in the scene.
\end{enumerate}
Both the removal and insertion paths are precomputed from the assembly geometry
and provided as input. Note that many algorithms exist to accomplish this~\cite{Bayraktar2026WAFR,assembleThemAll}.

The pseudocode for the generalized disassembly task is described in Alg. \ref{alg:generalized-disassembly-task}.
First, the current number of robot states (Line 1) and attachData
(Line 2) are stored in case a task fails and the initial state has to be restored.
This rollback and return failure behavior is repeated after every task that is run
(Line 4\textendash5, 9\textendash10, 16\textendash17, 20\textendash21, 24\textendash25).

The path to approach the object for the robot is calculated (Line 3).
After the robot plans to the grasp position, the attachData is stored (Line 6).
If the object has an associated removal path (Line 7), the robot follows the
path specified (Line 8). If the object has an insertion associated (Line 11), the desired end-effector
transformation is calculated from the start of the path and the relative transformation
between end-effector and object (Line 12) instead of the specified target (Line 14).
The robot then moves to the calculated position (Line 15).
If an insertion path is specified the robot follows the waypoints (Line 19).
After the goal transformation is achieved, the robot detaches the object (Line 22)
and leaves the area to make room for other robots (Line 23).
The algorithm signals success after all subtasks are completed (Line 26).

\subsection{Move Task}

The move task plans a collision-free path from the robot's current configuration
$q_\text{start} \in \mathcal{Q}$ to a specified goal pose
$T_\text{goal} \in \mathcal{W}_\text{free}$ in the
collision free workspace or a configuration $q_\text{goal} \in \mathcal{Q}$ in the
robot's configuration space. 

The pseudocode is given in Alg.~\ref{alg:move-task}. Given the robot's current
initial configuration (Line 1), the planner retries the movement request for $n_\text{retry}$ times (Line 2).
If a pose is specified as the goal, a set of goal configurations is generated
(Line 3), and the iteration is finished early if no goal states are feasible
(Line 4--5).
The planner is then queried to generate the motion (Line 6).
If the planner is successful in finding a solution (Line 7), the result is
committed to the robot's path storage (Line 8) and the task returns success
(Line 10). If all attempts are exhausted, the task returns false (Line 11).

\subsection{Attach and Detach Task}
These tasks represent the grasping of an object by the robot.
The attach task saves the homogeneous transformation between end-effector and
object at time $t_\text{attach}$.

\begin{equation}
  T_\text{EE}^\text{obj} = (T_\text{world}^\text{EE})^{-1} \cdot T_\text{world}^\text{obj}
\end{equation}

Saving the homogeneous transform between end-effector and object at the time of
execution allows for later reconstruction of the object position based on the robot
pose, as described in Sec. \ref{sec:collision-checking}.
This task does not move the robot to the object and solely tracks the relationship
between robot and object at a specific time.

The tracking is done through the \texttt{attachData}, which
resembles the association between robot and object at a specific time.

The detach task terminates the kinematic coupling by setting the time
$t_\text{detach}$ of the current attachData associated with the robot.

\begin{equation}
  T_\text{world}^\text{obj}(t) = T_\text{world}^\text{EE}(t) \cdot T_\text{EE}^\text{obj} \quad \forall \quad t_\text{attach} \le t \le t_\text{detach}
\end{equation}

Therefore, each attachData keeps track of:
\begin{itemize}
  \item $o_j$: A reference to an object
  \item $r_i$: A reference to a robot
  \item $t_\text{attach}$: The attach time
  \item $t_\text{detach}$: the detach time
  \item $T_\text{EE}^\text{obj}$: A relative transformation between object and end-effector of the robot
\end{itemize}

\subsection{Exit Task}
After placing an object, the robot has to vacate the area to avoid obstructing
subsequent operations by another robot. The exit task plans a path to a
location in a specified safe zone in the workspace, where it does not obstruct
any other robot.
The configuration is set to the home position of the robot
to guarantee a neutral starting position for the next task.

The goal used in the exit task is the final configuration until the robot is
reassigned a new task. If the configuration intersects with a committed trajectory
of another robot at any future time point, a collision occurs. Therefore, a
potential exit configuration has to be verified against all future time points
of other trajectories.

To check the candidate for validity, a linear sweep is performed from the start
of the exit task until the largest committed time point of any robot.

The pseudocode to check a configuration for future collisions is described in Alg. \ref{alg:check-future-collision}.
First, the lower time point of the time window is determined by the state passed to
check for future collision (Line 1). Then, the upper time is determined by largest
committed time point of any robot's trajectory (Line 2). While the time of the
state is smaller or equal to the maximum time point (Line 3), the state time is
increased incrementally by $\Delta t$ (Line 4 and 5). Each increment, the state of the
simulation is checked for collision (Line 6 and 7) and returns early on occurrence (Line 8).
After the complete time window is passed, the state is validated (Line 9).

With the introduced algorithms, the complete exit tasks can be formulated in
Alg. \ref{alg:exit-task}.
The robot considers the current position as the start position for the exit
task (Line 1). It tries up to a specified number of times (Line 2) to find
an exit plan. It first starts with an empty set of possible goal states (Line 3).
This set is filled with the specified number of valid exit configurations (Line 4).
First a candidate is drawn from the specified exit region (Line 5). Since the
robot might stay at the configuration until the simulation ends, the configuration
has to be checked for future collisions (Line 6). After the candidate is verified,
it is added to the possible goal states (Line 7).
If no goal states could be generated a retry is triggered early (Line 8 and 9).
The planner then tries to connect the start configuration to one of the possible
exit configurations (Line 10). If it succeeds (Line 11), the exit path is committed to
the planner's trajectory (Line 12) and returns successfully (Line 13). When all
retries are exhausted, the exit task returns failure (Line 14).

\input{src/algorithms/task_move}
\input{src/algorithms/task_object_extraction}

\input{src/algorithms/task_exit}

\subsection{Object Extraction Task\label{sec:task-object-extraction}}
Objects that are geometrically interlocked with adjacent parts are hard to move
freely through the workspace after grasping. Instead, they must follow a
precomputed path consisting of a sequence of waypoint transforms
$T_1, \ldots, T_m \in \text{SE}(3)$. These waypoints describe a collision-free
trajectory through the surrounding geometry. This procedure is used both for
extraction (removing an object from surroundings) and insertion (guiding an
object into its goal position).

The follow path task advances through the waypoints using inverse kinematics.
For each waypoint $T_i$, the required end-effector pose is
computed from the waypoint and the current grasp transform:
\begin{equation}
  T^\text{EE}_i = T_i \cdot (T^\text{obj}_\text{EE})^{-1}
\end{equation}
The IK solver attempts to find a joint configuration $\mathbf{q}_i$ that
realizes this end-effector pose.

The inverse kinematics problem for each waypoint can be stated as a
constrained optimization. The full space-time waypoint problem is:
\begin{equation}\label{eq:object-extraction-opt}
  \begin{aligned}
    \min_{\mathbf{q} \in \mathcal{Q},\;  t \in \mathbb{R}} \; \alpha t + (1-\alpha) \Vert q_i - q_{i-1} \Vert ^2
    \quad\text{s.t.} \quad
    T_\text{FK}(\mathbf{q}) &= T^\text{EE}_i \\
    \mathbf{q}_l \leq &\mathbf{q} \leq \mathbf{q}_u \\
    t > &t_{i-1} + \epsilon \\
    (\mathbf{q}, t) &\in \mathcal{Q}_\text{free}(t)
  \end{aligned}
\end{equation}
where $\mathcal{Q}_\text{free}(t)$ denotes the free configuration space at
time $t$, determined by the committed trajectories of all other robots and
objects.

Solving Eq. \ref{eq:object-extraction-opt} jointly is intractable in practice, as the
collision constraint couples the kinematic and temporal variables through the
full simulation state. The problem is therefore decomposed into two sequential
subproblems. First, the kinematic subproblem is solved independently
\begin{equation}
  \min_{\mathbf{q} \in \mathcal{Q}} \;
  \| T_\text{FK}(\mathbf{q}) - T^\text{EE}_i \|^2 + \lambda \Vert q_i - q_{i-1} \Vert ^2
  \quad \text{s.t.} \quad
  \mathbf{q}_l \leq \mathbf{q} \leq \mathbf{q}_u
\end{equation}
Second, the temporal subproblem finds the earliest
collision-free time for the IK solution:
\begin{equation}
  t_i = \min \{ t \geq t_{i-1} + \epsilon \mid
    (\mathbf{q}_i, t) \in \mathcal{Q}_\text{free}(t), \;
    t - t_{i-1} < t_\text{max} \}
\end{equation}
by incrementally advancing in steps of $\Delta t$ until a valid time is found
or the maximum time increase $t_\text{max}$ is exceeded. This decomposition
sacrifices joint optimality for tractability. Configurations can exist that would
be valid at an earlier time with a different $\mathbf{q}$ but are not
discoverable by the sequential approach. In practice, the waypoint density of
the precomputed paths is sufficient to ensure that the decoupled solution
reliably finds valid states.

%

The pseudocode for object extraction is described in Alg. \ref{alg:object-extraction}.
The algorithm starts by getting the last committed trajectory time point of
the robot (Line 1). It then iterates through each waypoint provided (Line 2).
First it calculates the end-effector transformation from the desired object path
(Line 3). It then solves the inverse kinematics problem for that transformation
(Line 5) to find the next configuration. If no valid configuration is found (Line 6),
a regrasp can be attempted if allowed (Line 7\textendash9). When the object does not
permit regrasps, a failure is returned (Line 8). After the regrasp is successful,
the desired end-effector transformation is updated with the new relative transformation
between end-effector and object (Line 10). This procedure is repeated until
a valid configuration is found (Line 11).

After a possible configuration is found, the last committed trajectory time point
is stored (Line 12). This time point is then used to iterate over a time window,
which is limited by an upper bound (Line 13). The solution from the inverse kinematics
is set to the relevant time point (Line 14) and validated for collision (Line 15).
If no collision occurs, the state is committed to the robot's trajectory (Line 16)
and continues to the next inverse kinematics problem (Line 17). In case of collision,
the time point is increased by a time delta $\Delta t$ (Line 19), until either a
valid time point is found or the time window is exhausted.

After the time window is over (Line 20) and no solution was found, a regrasp
is performed (Line 23) and the current waypoint is retried again (Line 24).
If no regrasp is permitted, the planner instead fails (Line 22) after no valid
solution was found.

When all waypoints find a suitable configuration and timepoint, the algorithm returns
success (Line 25).

%

%% file: src/algorithms/task_generalized_disassembly.tex
\begin{algorithm}[t]
  \caption{Generalized Disassembly Task}\label{alg:generalized-disassembly-task}
  \KwData{robot, object, $T_\text{goal}$, $t_\text{pick}$, $t_\text{place}$}
  \KwResult{success}
  $n_s \gets |\text{robot.states}|$ \tcp*{Rollback checkpoint}
  $n_a \gets |\text{object.attachData}|$\\
  \If{\textbf{not} \texttt{moveToObject(robot, object, $t_\text{pick}$)}}{
    \texttt{rollback($n_s$, $n_a$)} \\
    \Return failure}
  \texttt{attach(robot, object)}\\
  \If{object.removalPath $\neq \varnothing$}{
    \If{\textbf{not} \texttt{followPath(robot, object.removalPath)}}{
      \texttt{rollback($n_s$, $n_a$)} \\
      \Return failure}
  }
  \eIf{object.insertionPath $\neq \varnothing$}{
    $T_\text{goal} \gets T'_1 \cdot (T^\text{obj}_\text{EE})^{-1}$\\
  }{
    $T_\text{goal} \gets T_\text{goal} \cdot (T^\text{obj}_\text{EE})^{-1}$\\
  }
  \If{\textbf{not} \texttt{moveToGoal(robot, $T_\text{goal}$, $t_\text{place}$)}}{
    \texttt{rollback($n_s$, $n_a$)} \\
    \Return failure
  }
  \If{object.insertionPath $\neq \varnothing$}{
    \If{\textbf{not} \texttt{followPath(robot, object.insertionPath)}}{
      \texttt{rollback($n_s$, $n_a$)} \\
      \Return failure
    }
  }
  \texttt{detach(robot, object)}\\
  \If{\textbf{not} \texttt{exit(robot)}}{
    \texttt{rollback($n_s$, $n_a$)} \\
    \Return failure}
  \Return success
\end{algorithm}

%% file: src/algorithms/task_move.tex
\begin{algorithm}[t]
  \caption{Move Task}\label{alg:move-task}
  \KwData{robot, goal, $t_\text{I}$}
  \KwParam{$n_\text{retry}$}
  \KwResult{success}
  $\mathbf{q}_\text{I} \gets$ \texttt{getConfiguration(robot)}\\
  \For{$i \gets 1$ \KwTo $n_\text{retry}$}{
    goals $\gets$ \texttt{generateGoalStates(robot, goal)}\\
    \If{goals $= \varnothing$}{continue}
    path $\gets$ \texttt{plan(robot, $\mathbf{q}_\text{I}$, $t_\text{I}$, goals)}\\
    \If{path $\neq \varnothing$}{
      \texttt{appendPath(robot, path)}\\
      \Return true \\
    }
  }
  \Return false
\end{algorithm}

%% file: src/algorithms/task_object_extraction.tex
\begin{algorithm}[t]
  \caption{Object Extraction}\label{alg:object-extraction}
  \KwData{robot, object, path $= [T_1, \ldots, T_m]$}
  \KwParam{$t_\text{max}$, $\Delta t$}
  \KwResult{success}
  $t \gets$ \texttt{getLastStateTime(robot)}\\
  \For{$T_i \in$ path}{
    $T^\text{EE}_i \gets T_i \cdot (T^\text{obj}_\text{EE})^{-1}$\\
    \tcp{Solve IK, regrasping on failure}
    \Repeat{$\mathbf{q}_i \neq \varnothing$}{
      $\mathbf{q}_i \gets$ \texttt{solveIK(robot, $T^\text{EE}_i$)}\\
      \If{$\mathbf{q}_i = \varnothing$}{
        \If{\textbf{not} object.allowRegrasp}{\Return failure}
        \texttt{regrasp(robot, object, $t$)}\\
        $T^\text{EE}_i \gets T_i \cdot (T^\text{obj}_\text{EE})^{-1}$
          \tcp*{Recompute with new grasp}
      }
    }
    \tcp{Find earliest valid time}
    $t_\text{start} \gets t$\\
    \While{$t - t_\text{start} < t_\text{max}$}{
      $s \gets$ \texttt{createState($\mathbf{q}_i$, $t$)}\\
      \eIf{\texttt{isValid(s)}}{
        \texttt{robot.states.append($s$)}\\
        \textbf{break}
      }{
        $t \gets t + \Delta t$
      }
    }
    \If{$t - t_\text{start} \geq t_\text{max}$}{
      \If{\textbf{not} object.allowRegrasp}{\Return failure}
      \texttt{regrasp(robot, object, $t$)}\\
      \textbf{redo} current waypoint $T_i$
    }
  }
  \Return success
\end{algorithm}

%% file: src/algorithms/task_exit.tex
\begin{algorithm}[t]
  \caption{Exit Task}\label{alg:exit-task}
  \KwData{robot}
  \KwParam{$n_\text{retry}$, numberOfExitConfigurations}
  \KwResult{success}
  $\mathbf{q}_\text{start} \gets$ \texttt{getPose(robot)}\\
  \For{$i \gets 1$ \KwTo $n_\text{retry}$}{
    goalStates $\gets \varnothing$\\
    \For{$j \gets 1$ \KwTo numberOfExitConfigurations}{
      $\mathbf{q}_\text{exit} \gets$ \texttt{sampleExitRegion()}\\
      \If{\texttt{checkFutureCollision($q_\text{exit}$)}}{
	\texttt{append(goalStates, $\mathbf{q}_\text{exit}$)}\\
      }
    }
    \If{goalStates $= \varnothing$}{\textbf{continue}}
    path $\gets$ \texttt{plan(robot, $\mathbf{q}_\text{start}$, goalStates)}\\
    \If{path $\neq \varnothing$}{
      \texttt{appendPath(robot, path)}\\
      \Return true
    }
  }
  \Return false
\end{algorithm}

%% file: src/05_collisionchecking.tex
\section{Collision Checking and Grasp Generation}

Next we need to be able to generate grasp positions and do collision checking with existing robots and objects.

\subsection{Collision Checking\label{sec:collision-checking}}

All tasks rely on a collision checking algorithm, which checks if a certain configuration is in collision at
a specific time point.
The pseudocode for collision checking is described in Alg. \ref{alg:collision-checking}.
First, the associated time with the state is acquired (Line 1). Then, the simulation
state is set with Alg. \ref{alg:set-simulation-state} (Line 2). Then the new
configuration is set (Line 3).
Since it is unknown if an object is attached at that time point to the robot,
each object (Line 4) is checked if it is attached to the desired robot.
First, the relevant attachment data for that time point is searched with Alg. \ref{alg:find-attachment-data}
(Line 5). If the robot associated with the attachData matches the selected robot (Line 6),
the current end-effector pose is saved (Line 7) and combined with the saved relative
transform (Line 8) between object and robot from the attachData to the actual pose
of the object at that time point (Line 9). Then the pose of the object is set (Line 10)
and the loop exits early (Line 11).
After each object and robot is set, the algorithm returns if a collision is detected
by the collision detector (Line 12).
\input{src/algorithms/algorithm_collision_checking}
\input{src/algorithms/algorithm_find_attach_data}

\input{src/algorithms/algorithm_set_simulation_state}

\subsubsection{Find Attachment Data\label{sec:find-attachment-data}}
Since objects can be attached to different robots at different time points,
information about the attachment time, detachment time and relative transform between
end-effector and object has to be saved to restore the complete state at any time
point. Attachment data is stored in chronological order of occurrence.

\input{src/figures/figure_attachement_data}

The pseudocode to find the attachData at a specific time point is
described in Alg. \ref{alg:find-attachment-data}.
First, the associated data
storage is selected (Line 1). It then verifies that the object was actually moved
(Line 2). Since the data is stored in chronological order, the first attach time
is used as an early return to indicate that the object is still in its initial
pose (Line 3\textendash5). The same behavior is true for every time point after
the last detach time point. The last detach time (Line 6) is compared against the
timepoint (Line 7) to determine if an early return (Line 8) is feasible.
Both of the early returns return their attachData as a comparison against the timepoint
to allow to check if the object should be placed in the start or goal pose.

If neither condition is met, a linear sweep across the data (Line 9) is combined
with a range check on the time point with a time window, consisting of the
attach and detach time (Line 10). The relevant attachData is then finally returned
(Line 11).

\subsubsection{Setting the Simulation State\label{sec:set-simulation-state}}

To execute collision checking at a specific time point, every object and robot has to be moved to the correct configuration. We call this the setting of the simulation state. 
The pseudocode to set the simulation state is described in Alg. \ref{alg:set-simulation-state}.
First, the algorithm iterates over each object to set their pose (Line 1).
The relevant attachment data for each object is located (Line 2), as outlined in
Alg. \ref{alg:find-attachment-data}. The time step is categorized as either prior to,
within or subsequent to the attachment data (Line 4). If the time step is prior to the
first attachment data, the object has not yet been moved and remains at the starting position.
If the time step falls within the time period of an attachment, the time itself is utilized
to set the pose of the robot. Conversely, if the time step is larger than the last time period,
the detach time of the last attachment data is used for the pose.
The robot's states which lie right before and after the time point are identified (Line 5).
Then, the state is interpolated (Line 6) and the robot's configuration is set (Line 7) for the time point.
This is done, so the end-effector transformation can be acquired (Line 8).
This is used with the relative end-effector-object transformation (Line 9) to
reconstruct the object's world transformation (Line 10). The object is then set
to that pose (Line 11).

After each object is placed to the correct transformation, each robot can be set (Line 12).
Now the input time is used to determine the bounding states (Line 13)
and interpolation (Line 14) like before. Then the robot is set to the interpolated
configuration (Line 15).

\subsection{Grasp Pose Generation\label{sec:grasp-pose-generation}}
Each surface sample provides a target position for the inverse kinematic problem
of the end-effector. Since the omnidirectional gripper can approach from any orientation,
the orientation is not constrained during solving and a random orientation is
sampled for each IK query. The inverse kinematics solver uses different restarts with random
initial guesses sampled uniformly within the robot's configuration space
$\mathcal{Q}$.

The goal state generation pseudocode is described in Alg. \ref{alg:grasp-pose-generation}.
First, an empty goalStates set is initialized (Line 1).
Then, the algorithm tries up to a certain number of tries (Line 2) to
generate at most a specified number of possible goal states (Line 3) and return
early if the amount is met (Line 4).
A point on the object is sampled (Line 5) and used as the target position
for the inverse kinematics problem (Line 6). If no valid configuration can be
found (Line 7), the next iteration begins (Line 8).
When a valid configuration is found, it is only kinematically feasible and
has to be checked to be free in the future (Line 9).
After the state was validated it is added to the set of possible goal states
for the planner (Line 11).
The algorithm returns the set of all found goal state candidates (Line 12).

\input{src/algorithms/algorithm_goal_state_generation}

\subsubsection{Check Future Collisions\label{sec:check-future-collisions}}

Not every IK solution represents a valid grasp configuration. A candidate is only
accepted if it passes a future collision check with Alg.\,\ref{alg:check-future-collision}.
This prevents selecting grasp poses that are feasible but invalidated by
other robots' trajectories at different time points.

\input{src/figures/figure_timeline_future_collision}
\input{src/algorithms/check_future_collision}

\subsubsection{Surface Sampling\label{sec:surface-sampling}}

A grasp can occur on any point on the surface of the object since the end-effector
is modeled as an omnidirectional gripper, as described in the first assumption
in Sec. \ref{sec:problem-statement}. 
The grasp pose generation produces a set of possible candidates in $\mathcal{Q}$
for the motion planner. These candidates are generated by sampling contact points
on the surface of the object and then used as the target for inverse kinematics.

A point on the object surface serves as the target position for the
end-effector.

The object surface is considered to be described as a mesh. Each mesh
consists of vertices, edges and faces. Vertices describe positions in the
space. Edges are connections between vertices. A closed set of edges is called
a face. In this case, the faces are considered to be triangle faces, resulting
in a triangle mesh.

For mesh objects, a face is selected with probability proportional
to its area to ensure a uniform distribution of sample points across the
surface. The cumulative area distribution over all faces is computed and
a face is selected via inverse transform sampling.
\begin{equation}
    A_i = \frac{1}{2} \Vert (v_1 - v_0) \times (v_2 - v_0) \Vert
\end{equation}
The probability of selecting a face $i$ from a mesh with $N$ faces
\begin{equation}
  P(i) = \frac{A_i}{\sum_{j=1}^{N}A_j}
\end{equation}
Then a uniform sample from the distribution is used to select a face.

Within the selected
face, a point is sampled using barycentric coordinates with two uniform
random variables $r_1, r_2 \in [0, 1]$:
\begin{equation}
  \mathbf{p} = (1 - r_1) \mathbf{v}_0 + r_1 (1 - r_2) \mathbf{v}_1 + r_1 r_2 \mathbf{v}_2
\end{equation}
where $\mathbf{v}_0, \mathbf{v}_1, \mathbf{v}_2$ are the faces' vertices.
The result is transformed to world coordinates using the object's current pose.

%% file: src/algorithms/algorithm_collision_checking.tex
\begin{algorithm}[t]
  \caption{Collision Check}\label{alg:collision-checking}
  \KwData{$q_\text{new}$, $robot_\text{selected}$}
  \KwResult{boolean}
  $t_\text{at} \gets$ \texttt{getTime($q_\text{new}$)}\\
  \texttt{setSimulationState($t_\text{at}$)} \\

  \texttt{setConfiguration($\text{robot}_\text{selected}$, $q_\text{new}$)}\\
  \For{$o_j \in \mathcal{O}$ }{
    data $\gets$ \texttt{findAttachmentData($o_j$, $t_\text{at}$)}\\
    \If{$o_j.\text{robot} == \text{robot}_\text{selected}$}{
      eeTF $\gets$ robot.eeTF \\
      eeToObjTF $\gets$ data.eeToObjTF \\
      $objTF \gets eeTF * eeObjTF $\\
      \texttt{setPose(obj, objTF)}\\
      \textbf{break}\\
    }
  }
  \texttt{return isSimInCollision()}
\end{algorithm}

%% file: src/algorithms/algorithm_find_attach_data.tex
\begin{algorithm}[t]
  \caption{Find Attachment Data}\label{alg:find-attachment-data}
  \KwData{object, $t_\text{at}$}
  \KwResult{Attachment Data}
  attachData $\gets$ \texttt{getAssociatedAttachData(object)} \\
  \textbf{assert} attachData $\ne \varnothing$ \\
  firstAttachTime $\gets$ \texttt{attachData.front().attachTime}\\
  \If{ $t_\text{at} < $ firstAttachTime }{
    \texttt{return attachData.front()}}
  lastDetachTime $\gets$ \texttt{attachData.back().detachTime}\\
  \If{ $t_\text{at} > $ lastDetachTime}{
    \texttt{return attachData.back()}}
  \ForEach {$data \in attachData$}{
    \If{$t_\text{at} > t_\text{attach} \wedge t_\text{at} < t_\text{detach}$}{\texttt{return} data}
  }
\end{algorithm}

%% file: src/algorithms/algorithm_set_simulation_state.tex
\begin{algorithm}[t]
  \caption{Set Simulation State}\label{alg:set-simulation-state}
  \KwData{$t_\text{at}$}
  \For{obj $\in$ objects}{
    data $\gets$ \texttt{findAttachmentData(obj, $t_\text{at}$)}\\
    robot $\gets$ \texttt{getRobot(data)} \\
    time $\gets$ \texttt{determineTimePoint(data, $t_\text{at}$)} \\
    $q_\text{lower}$, $q_\text{upper}$ $\gets$ \texttt{findBoundingStates(robot, time)}\\
    $q_\text{inter} \gets $ \texttt{interpolate($q_\text{lower}$, $q_\text{upper}$, time)}\\
    \texttt{setConfiguration(robot, $q_\text{inter}$)}\\
    eeTF $\gets$ robot.eeTF \\
    eeToObjTF $\gets$ data.eeToObjTF \\
    $objTF \gets eeTF * eeObjTF $\\
    \texttt{setPose(obj, objTF)}\\
  }
  \For{robot $\in$ robots}{
    $q_1$, $q_2$ $\gets$ \texttt{findBoundingStates(robot, $t_\text{at}$)}\\
    $q_\text{inter} \gets $ \texttt{interpolate($q_1$, $q_2$, time)}\\
    \texttt{setConfiguration(robot, $q_\text{inter}$)}\\
  }
\end{algorithm}

%% file: src/figures/figure_attachement_data.tex
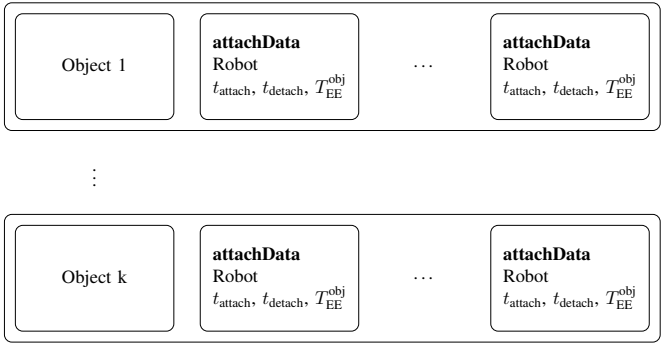
\begin{figure}[t]
\begin{center}
\resizebox{\linewidth}{!}{
  \begin{tikzpicture}
    \node[draw, rounded corners, minimum width=3cm, minimum height=2cm, align=left] (obj1) {Object 1};
    \node[draw, rounded corners, minimum width=3cm, minimum height=2cm, align=left, right of= obj1, node distance=3.5cm] (data11) {\textbf{attachData}\\ Robot \\ $t_\text{attach}$, $t_\text{detach}$, $T_\text{EE}^\text{obj}$};
    \node[draw, rounded corners, minimum width=3cm, minimum height=2cm, align=left, right of= data11, node distance=5.5cm] (data12) {\textbf{attachData}\\ Robot \\ $t_\text{attach}$, $t_\text{detach}$, $T_\text{EE}^\text{obj}$};
    \node[draw, rounded corners, fit=(obj1) (data11) (data12), inner sep=2mm] (Outer) {};
    \path (data11.east) -- (data12.west) node[midway] {\dots};

    \node[draw=none, align=left, below of= obj1, node distance=2cm] (dotsleft) {\vdots};

    \node[draw, rounded corners, minimum width=3cm, minimum height=2cm, align=left, below of= dotsleft, node distance=2.0cm] (obj2) {Object k};
    \node[draw, rounded corners, minimum width=3cm, minimum height=2cm, align=left, right of= obj2, node distance=3.5cm] (data21) {\textbf{attachData}\\ Robot \\ $t_\text{attach}$, $t_\text{detach}$, $T_\text{EE}^\text{obj}$};
    \node[draw, rounded corners, minimum width=3cm, minimum height=2cm, align=left, right of= data21, node distance=5.5cm] (data22) {\textbf{attachData}\\ Robot \\ $t_\text{attach}$, $t_\text{detach}$, $T_\text{EE}^\text{obj}$};
    \node[draw, rounded corners, fit=(obj2) (data22), inner sep=2mm] (Outer) {};
    \path (data21.east) -- (data22.west) node[midway] {\dots};

  \end{tikzpicture}
}
\end{center}
\caption{Storage representation of attachment data}
\end{figure}

%% file: src/algorithms/algorithm_goal_state_generation.tex
\begin{algorithm}[t]
  \caption{Grasp Pose Generation}\label{alg:grasp-pose-generation}
  \KwData{robot, object}
  \KwParam{maxTryNumber, maxNumGoals}
  \KwResult{goalStates}
  goalStates $\gets \varnothing$\\
  \For{$i \gets 0$ \KwTo maxTryNumber}{
    \If{$|$goalStates$| \geq $ maxNumGoals}{
      \Return goalStates
    }
    surfacePoint $\gets$ \texttt{sampleSurface(object)}\\
    q $\gets$ \texttt{solveIK(robot, surfacePoint)}\\
    \If{$q = \varnothing$}{continue}
    \If{not \texttt{checkFutureCollision(q)}}{continue}
    \texttt{append(goalStates, q)} \\
  }
  \Return goalStates
\end{algorithm}

%% file: src/figures/figure_timeline_future_collision.tex
\begin{figure}[t]
\centering
\begin{tikzpicture}[
    x=\linewidth/17,    
    block/.style={draw, thick, minimum height=0.8cm, anchor=west, inner sep=2pt,
                  font=\footnotesize, align=center},
    r1/.style={block, fill=blue!20},
    r2/.style={block, fill=red!20},
    event/.style={very thick, font=\scriptsize},
    rlabel/.style={font=\small\bfseries, anchor=east, black},
    time_tick/.style={font=\small, black},  
]
\node[rlabel] at (-0.4, 0) {$A$};
\node[r1, minimum width=4*(\linewidth/17)] at (0, 0) {Move};
\draw[event, blue!60] (4, -0.4) -- (4, 0.4);
\node[event, blue!60, above] at (4, 0.45) {Pick};
\node[r1, minimum width=7*(\linewidth/17)] at (4, 0) {Move};
\draw[event, blue!60] (11, -0.4) -- (11, 0.4);
\node[event, blue!60, above] at (11, 0.45) {Place};
\node[r1, minimum width=3*(\linewidth/17), fill=blue!35] at (11, 0) {Exit};

\node[rlabel] at (-0.4, -1.5) {$B$};
\node[r2, minimum width=3*(\linewidth/17)] at (0, -1.5) {Move};
\draw[event, red!60] (3, -1.9) -- (3, -1.1);
\node[event, red!60, above] at (3, -1.05) {Pick};
\node[r2, minimum width=4*(\linewidth/17)] at (3, -1.5) {Move};
\draw[event, red!60] (7, -1.9) -- (7, -1.1);
\node[event, red!60, above] at (7, -1.05) {Place};
\node[r2, minimum width=2*(\linewidth/17), fill=red!35] at (7, -1.5) {Exit};

\node[block, minimum width=5*(\linewidth/17), fill=red!10, draw=red!40, densely dashed]
    at (9, -1.5) {Stationary};

\draw[red, very thick] (9, -2.4) -- (9, -2.6) -- (14, -2.6) -- (14, -2.4);
\draw[red, very thick] (11.5, -2.6) -- (11.5, -2.8);
\node[font=\footnotesize, red, align=center, below] at (11.5, -2.8)
    {Collision can occur};

\draw[red, thick, dashed] (9, 0.9) -- (9, -2.6);
\node[time_tick, above] at (9, 0.9) {$t_{\text{exit},B}$};

\draw[->, thick, gray] (0, -3.8) -- (15, -3.8);
\node[font=\small, black, anchor=west] at (15, -3.8) {$t$};
\draw[gray] (0, -3.7) -- (0, -3.9);
\node[time_tick, below] at (0, -3.9) {$0$};
\draw[gray] (9, -3.7) -- (9, -3.9);
\node[time_tick, below] at (9, -3.9) {$t_{\text{exit},B}$};
\draw[gray] (14, -3.7) -- (14, -3.9);
\node[time_tick, below] at (14, -3.9) {$t_{\text{exit},A}$};
\end{tikzpicture}
\caption{Timeline illustration for future collision check}
\label{fig:future-collision-timeline}
\end{figure}
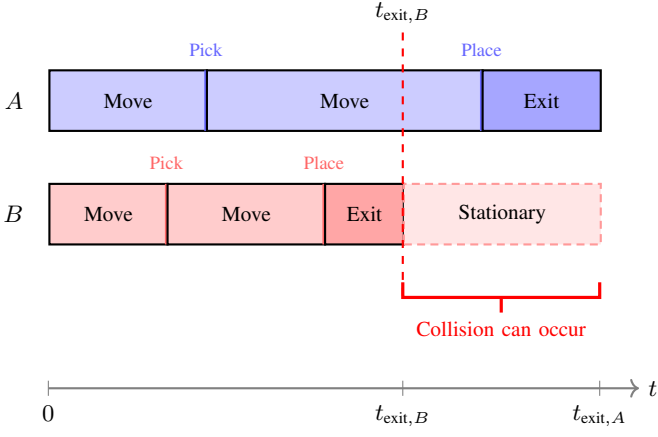

%% file: src/algorithms/check_future_collision.tex
\begin{algorithm}[t]
  \caption{Check Future Collision}\label{alg:check-future-collision}
  \KwData{robot, $\mathbf{q}$, $t_\text{at}$}
  \KwParam{$\Delta t$}
  \KwResult{validConfiguration}
  $t \gets t_\text{at}$\\
  $t_\text{max} \gets \texttt{getMaxTrajectoryTime()}$\\
  \While{$t \leq t_\text{max}$}{
    $t \gets t + \Delta t$\\
    \texttt{setSimTime(t)} \\
    \texttt{setConfiguration(robot, q)} \\
    \If{\texttt{isInCollision()}}{
      \Return false
    }
  }
  \Return true
\end{algorithm}

%% file: src/06_experiments.tex
\section{Evaluation}
\input{src/figures/scenarios}
\input{src/figures/results_makespan_and_calculation}
\input{src/figures/results_pareto_front}
\input{src/figures/results_scheduling}
\input{src/figures/results_failure_analysis}

To evaluate our disassembly planner, we perform an evaluation on six scenarios
shown in Fig.~\ref{fig:scenarios}. For each scenario, we vary the number of
robots from 1 to 9 to showcase the robustness of the planner to different team
sizes and to demonstrate the optimal number of robots required for a specific
scenario. Finally, we compare ST-RRT* as individual planner to two alternatives,
namely RRT* with 5s and 10s time bounds, respectively.

\subsection{Implementation Details}

CoMuDi is implemented using the
Open Motion Planning Library (OMPL) \cite{sucan2012the-open-motion-planning-library}.
Both RRT* \cite{karaman2011sampling} and ST-RRT* \cite{STRRTstar} are used with
an optimization objective to minimize the arrival time. The arrival time
is the last time point of a trajectory for one planner query and is not the
same as the makespan, which is the total time needed to complete the disassembly
process, introduced in the problem statement in  Sec. \ref{sec:problem-statement}.
A maximum computation time $t_\text{timelimit}$ of 10s
is used as a termination condition for the planner. After a solution is
found shortcutting and B-spline smoothing are applied to the path.

\subsection{Parameters}

The tunable parameters of CoMuDi are set to the following values. The maximum number of surface-sampling and inverse-kinematics attempts during grasp pose generation (Alg.~\ref{alg:grasp-pose-generation}) is set to \textbf{maxTryNumber}$=1000$. The maximum number of goal configurations collected before grasp pose generation returns (Alg.~\ref{alg:grasp-pose-generation}) is set to \textbf{maxNumGoals}$=5$.
The time increment for forward collision sweeps in future-collision checking (Alg.~\ref{alg:check-future-collision}) is set to $\Delta t = 0.1$.
For planning, we use $n_\text{retry}=3$ for maximum number of planning attempts for the move and exit task. Further, for the exit task, we set \textbf{numberOfExitConfigurations} to 10. For the object extraction task, we set $t_\text{max}=10$. For the IK solver, we use $\epsilon=0.1$ (Minimum time separation required between consecutive waypoint states) and $\lambda=1$ (Weight on joint-displacement regularization).

\subsection{Benchmark Hardware}

All runs are performed on a Dell XPS 15 9530 laptop, equipped with an 
Intel Core i7-13700H processor. 
Each trial is pinned on performance core 0 of the CPU with a stable clock 
speed between 4.7 and 4.8 GHz under sustained load. 
Runs were executed sequentially since the CPU frequency varied
for each core under load if multiple instances ran in parallel. This ensures 
comparable results within the benchmarks.

\subsection{Multi-Robot Disassembly Benchmark}

For each scenario, we report on computation time, makespan (total time elapsed from start
to finish), Pareto front (makespan versus computation
time), scheduling timeline, and analyze the failure rate.

\subsubsection{Computation Time and Makespan}

The first metric we evaluate is computation time and makespan. 
The computation time captures the cumulative time 
consumed across all queries needed to create the sequence, while the makespan captures the total time required to complete the planned
disassembly sequence. The results are shown in
Fig.~\ref{fig:calculation-time-and-makespan}.
For both plots, we report the median with the shaded bands denoting the interquartile
range\footnote{The interquartile range marks the area in which 50\% of the
samples fall.} across runs. 
The makespan is expected to decrease, while the execution time per task is expected 
to grow as individual assignments take place in an increasingly populated
workspace. It can be observed that this remains true for all six scenarios, with
makespan decreasing and computation time increasing.

\subsubsection{Pareto Front}

While minimizing makespan is a high priority for disassembly tasks, it comes at
the cost of increasing computation time. To better understand this trade-off, we
visualize those two metrics in a Pareto front plot in
Fig.~\ref{fig:pareto-front}.
The plot shows one sample per robot count, with the number of robots written
inside. A blue line connects all samples which are lying on the pareto frontier,
meaning that no other sample is dominating it. A sample dominates another
sample, if both makespan and computation time are lower. It can be observed that
all samples are on or are lying close to the pareto frontier, meaning that varying the
number of robots clearly trades-off makespan and computation time. We can also
observe that a number of 3 to 5 robots is often a good compromise between those
two metrics.

\subsubsection{Scheduling Timeline}

Minimizing makespan is often influenced by the idle time of the robots, with
less idle time often being preferable. We visualize this metric for all 9 robots in
Fig.~\ref{fig:scheduling-timeline}. For each robot, we visualize how much time
is spent being idle (white), executing a pick move (blue), a place move
(orange), a pull move (red), or an exit move (brown).
It can be seen that idle time is minimized except when the nature of the
scenario prevents this. For example the crate scenario is highly parallel, in
that robots have no tasks left once they solve their assigned task.

\subsubsection{Failure Rate Analysis\label{sec:results-failure-analysis}}

Finally, we conduct a failure rate analysis as shown in
Fig.~\ref{fig:results-failure-rates}. This analysis gives a more detailed
insight into our planner, and shows how failures are changing with the number
of robots. In detail, we plot the percentage of success (green) for each robot
count, together with the percentages of four failure modes: exit mode failure (light
green), pull mode failure (yellow), plan to object failure (blue) and plan to goal failure (purple). Those failure
modes do not stop the algorithm, but they represent wasted computation time,
which we want to minimize for better computation time.

\input{src/figures/results_rrtstar_comparison.tex}
\subsection{Individual Planner Comparison\label{sec:rrtstar-comparison}\label{sec:tower-problem}}

This section compares RRT* and ST-RRT* in the context of the tower scenario. We compare makespan and success rate while varying the number of robots from one to nine.
Each run has a timeout of 1000 seconds, 10 runs per robot count, and an
individual planner timeout of 10 seconds.

Since RRT* can only operate within a fixed time space~\cite{STRRTstar}, we
evaluate two time
windows. A 5 second window is used as a tight bound with little
room for detours, but easier to optimize due to the smaller state space.
A 10 second window allows for greater flexibility in space-time,
enabling detours to be taken to avoid robots. 
Both time windows were empirically selected based on the convergence behavior of
ST-RRT* in Fig.~\ref{fig:calculation-time-and-makespan}. These bounds are otherwise not known a priori.

The scaling of the makespan for all three methods is shown in
Fig.~\ref{fig:rrtstar-compare-scaling}.
With a 5 second time window, the makespan for RRT* is the highest at the start, with
168s for one robot, 82s for three robots, until it closes the gap with 26s for
five robots. After five robots, the makespan matches or outperforms that of the ten-second time window.
With the 10s time window, the performance is similar to ST-RRT* at the
beginning, from 99 seconds for one robot to 27 seconds for five robots. Afterwards, the makespan increases
again, fluctuating between 31 and 36 seconds for six to nine robots.
ST-RRT* consistently outperforms the makespan of the other two methods. It starts
at 94 seconds with one robot and falls monotonically to 20 seconds with nine robots.

The success rate of the three methods is shown in
Fig.~\ref{fig:rrtstar-compare-success-rate}.
RRT* with a 10s window starts at 100\% for one robot, but drops rapidly
to around 30\% for three or more robots. The 5s window for RRT*
has a success rate of around 50\%, dropping to 20\% with four and six robots. ST-RRT* achieves a success rate of
100\% throughout.

%% file: src/figures/scenarios.tex
\newcommand{\scenarioheight}{4.2cm}
\newcommand{\scenariowidth}{0.32\textwidth}

\begin{figure*}[t]
    \centering
    
    \begin{subfigure}[b]{\scenariowidth}
        \centering
        \includegraphics[height=\scenarioheight,width=\textwidth,keepaspectratio]{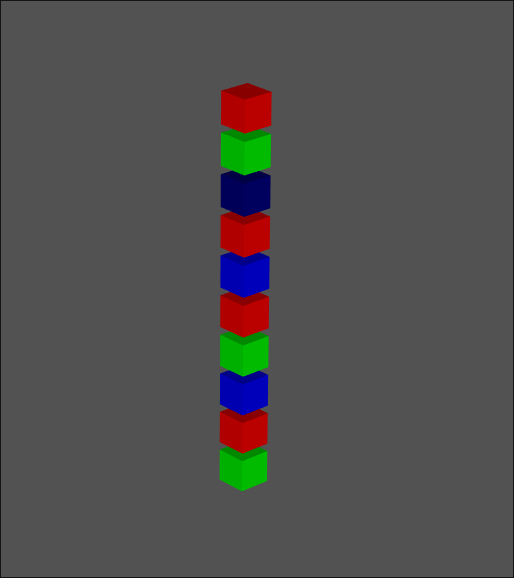}
        \caption{Tower (10 Components)}
        \label{fig:scenario-tower}
    \end{subfigure}
    \begin{subfigure}[b]{\scenariowidth}
        \centering
        \includegraphics[height=\scenarioheight,width=\textwidth,keepaspectratio]{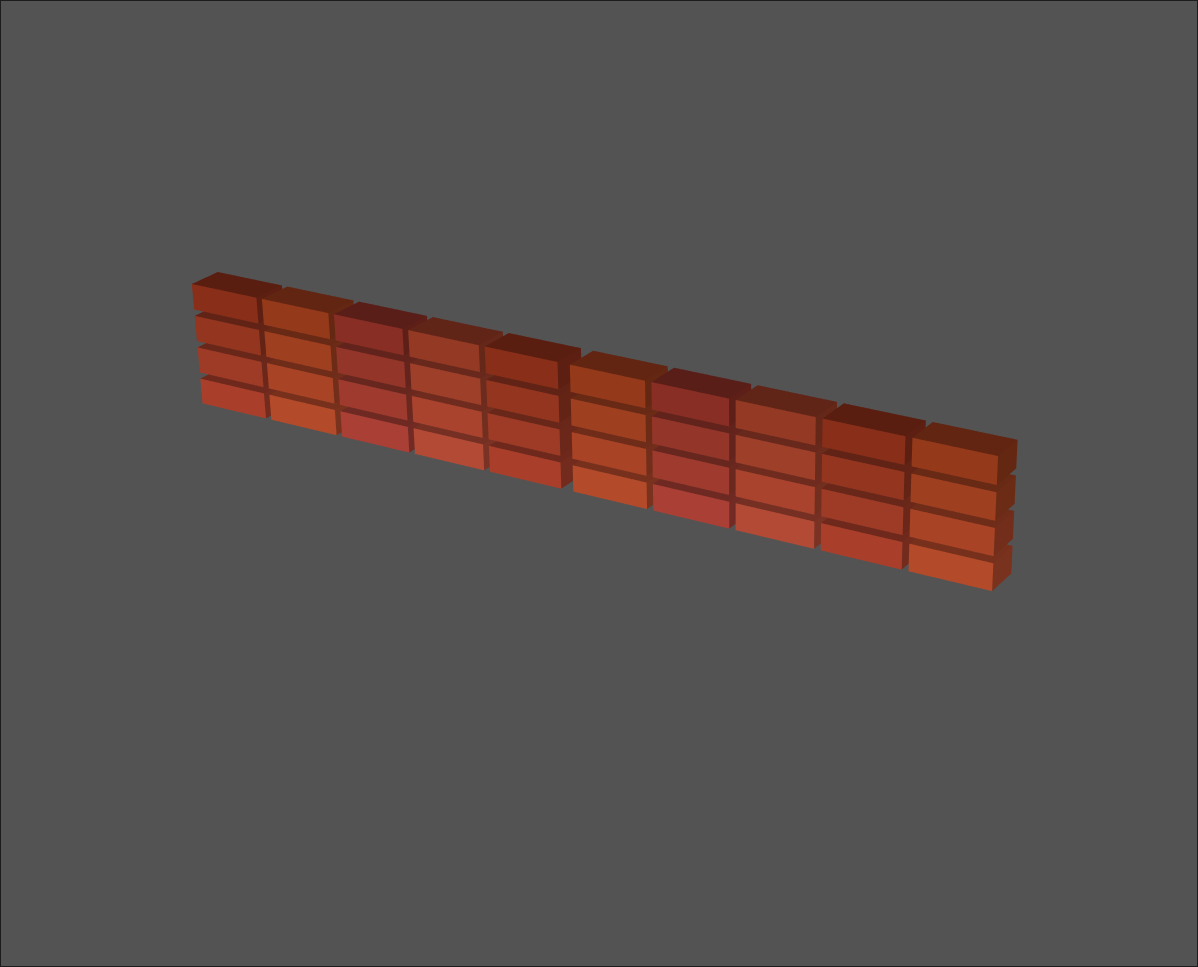}
        \caption{Wall (40 Components)}
        \label{fig:scenario-wall}
    \end{subfigure}
    \begin{subfigure}[b]{\scenariowidth}
        \centering
        \includegraphics[height=\scenarioheight,width=\textwidth,keepaspectratio]{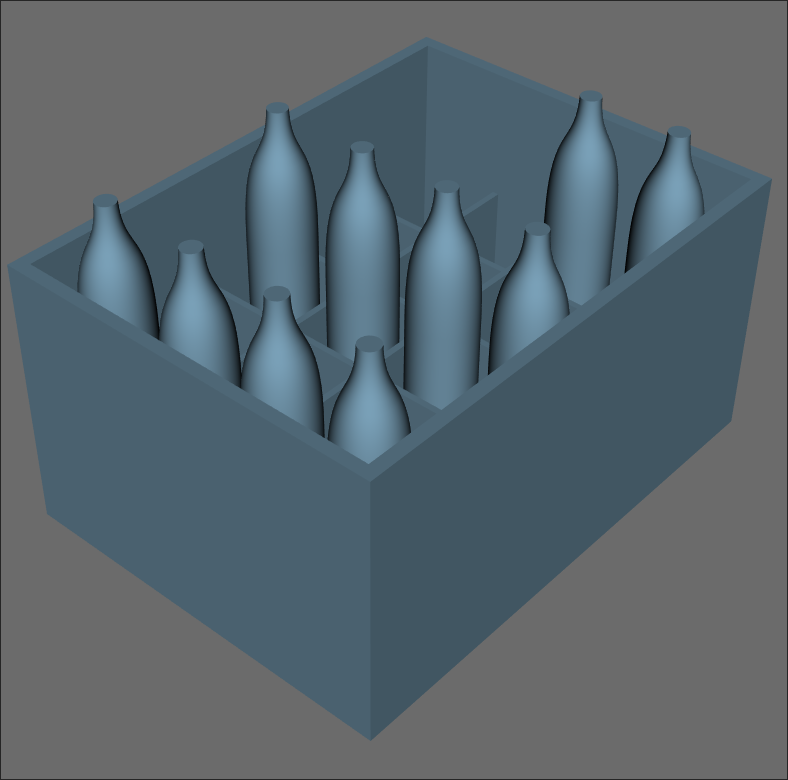}
        \caption{Crate (10 Components)}
        \label{fig:scenario-crate}
    \end{subfigure}
    
    \vspace{1.2em}
    
    \begin{subfigure}[b]{\scenariowidth}
        \centering
        \includegraphics[height=\scenarioheight,width=\textwidth,keepaspectratio]{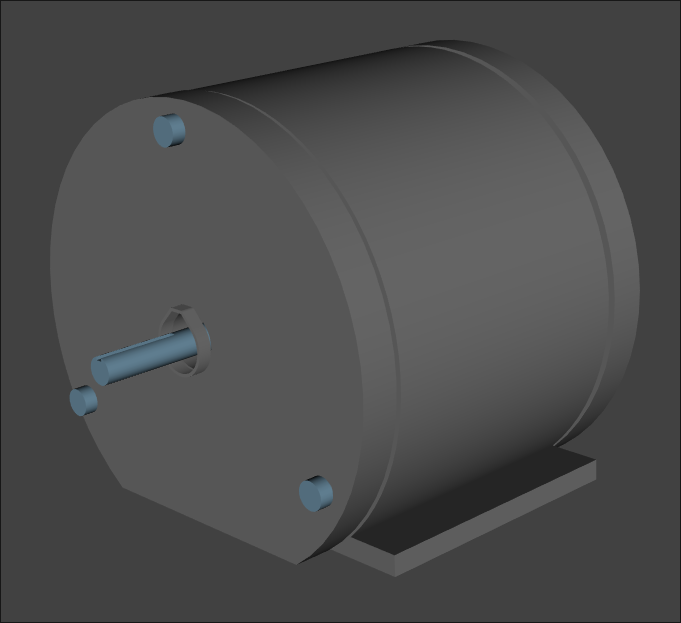}
        \caption{Motor (13 Components)}
        \label{fig:scenario-motor}
    \end{subfigure}
    \begin{subfigure}[b]{\scenariowidth}
        \centering
        \includegraphics[height=\scenarioheight,width=\textwidth,keepaspectratio]{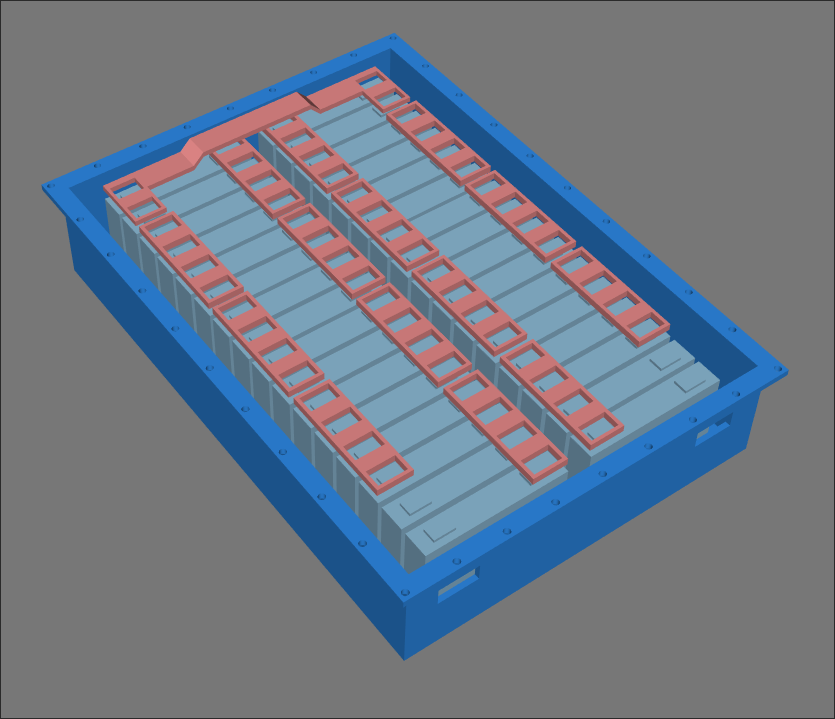}
        \caption{Battery (49 Components)}
        \label{fig:scenario-battery}
    \end{subfigure}
    \begin{subfigure}[b]{\scenariowidth}
        \centering
        \includegraphics[height=\scenarioheight,width=\textwidth,keepaspectratio]{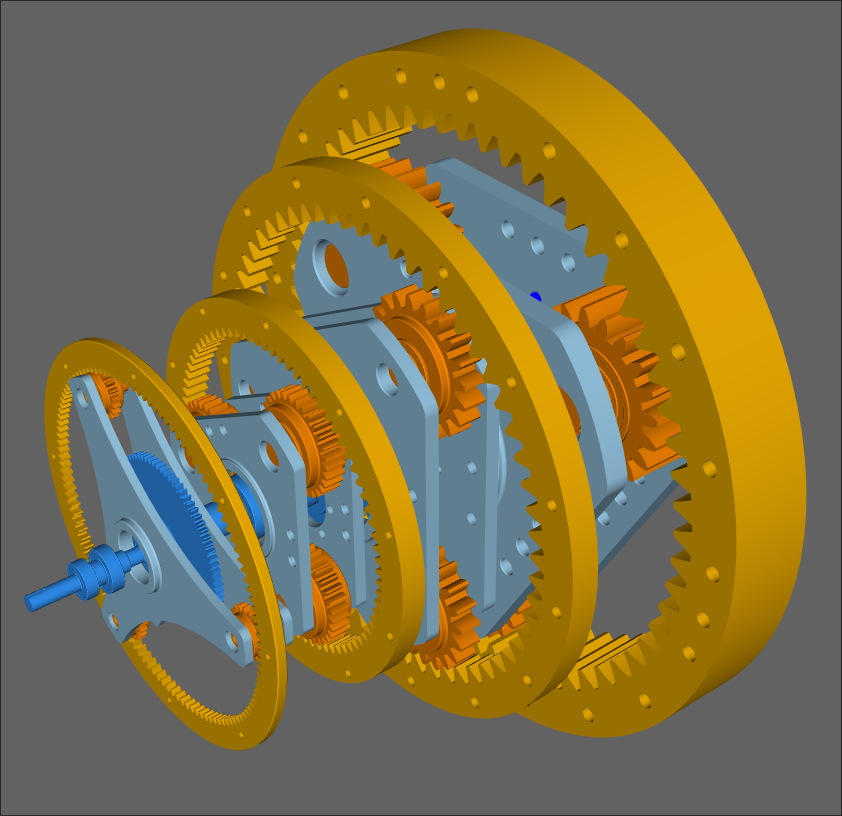}
        \caption{Gearbox (33 Components)}
        \label{fig:scenario-gearbox}
    \end{subfigure}
    
    \caption{Scenarios used in the disassembly experiments.\label{fig:scenarios}}
\end{figure*}

%% file: src/figures/results_makespan_and_calculation.tex
\newcommand{\scalingheight}{4.8cm}     
\newcommand{\scalingwidth}{0.48\textwidth}   

\begin{figure*}[t]
    \centering
    
    \begin{subfigure}[b]{\scalingwidth}
        \centering
        \includegraphics[height=\scalingheight,width=0.49\textwidth,keepaspectratio]{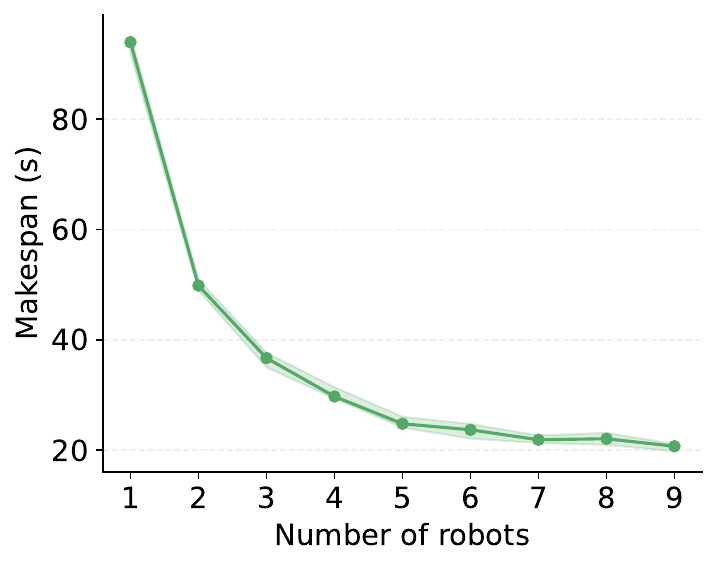}
        \includegraphics[height=\scalingheight,width=0.49\textwidth,keepaspectratio]{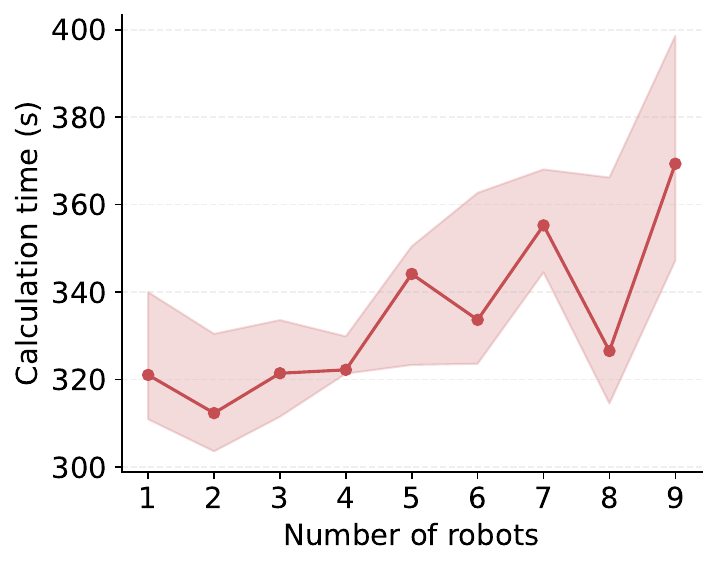}
        \caption{Makespan and Computation Time for Tower}
        \label{fig:scaling-tower}
    \end{subfigure}
    \hfill
    \begin{subfigure}[b]{\scalingwidth}
        \centering
        \includegraphics[height=\scalingheight,width=0.49\textwidth,keepaspectratio]{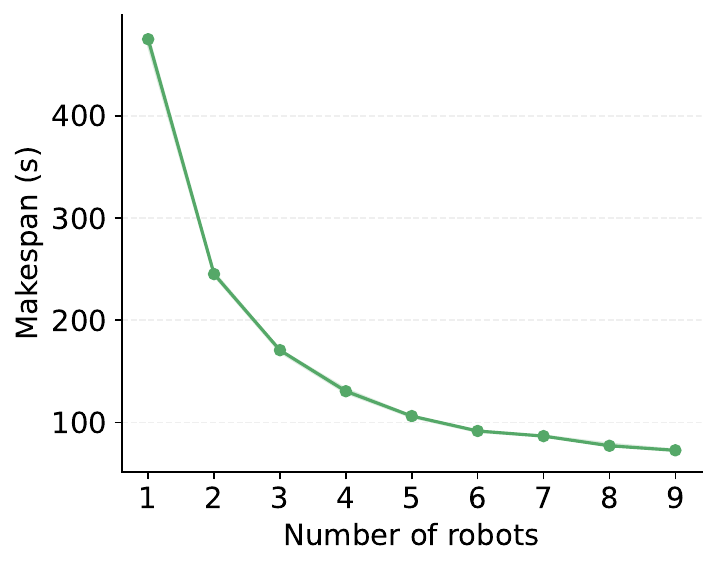}
        \includegraphics[height=\scalingheight,width=0.49\textwidth,keepaspectratio]{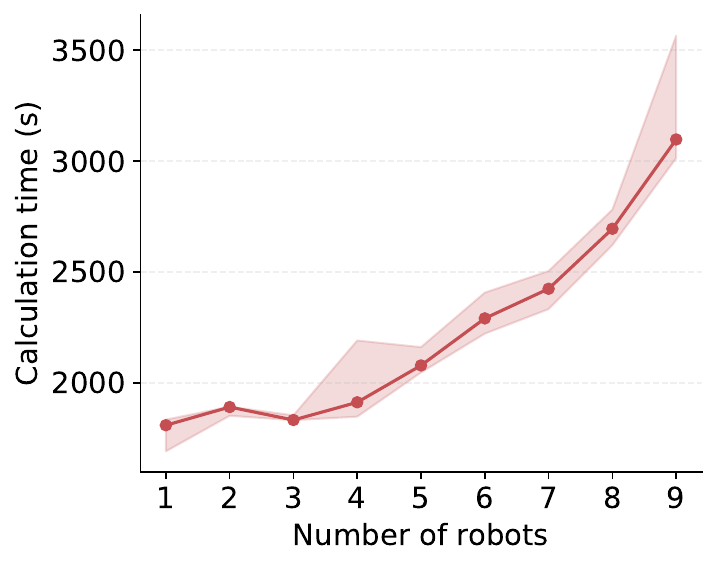}
        \caption{Makespan and Computation Time for Wall}
        \label{fig:scaling-wall}
    \end{subfigure}
    
    \vspace{0.8em}
    
    \begin{subfigure}[b]{\scalingwidth}
        \centering
        \includegraphics[height=\scalingheight,width=0.49\textwidth,keepaspectratio]{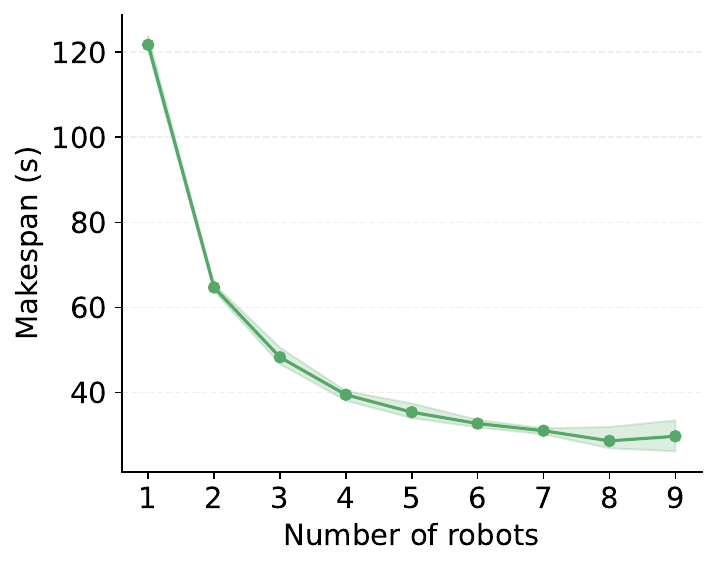}
        \includegraphics[height=\scalingheight,width=0.49\textwidth,keepaspectratio]{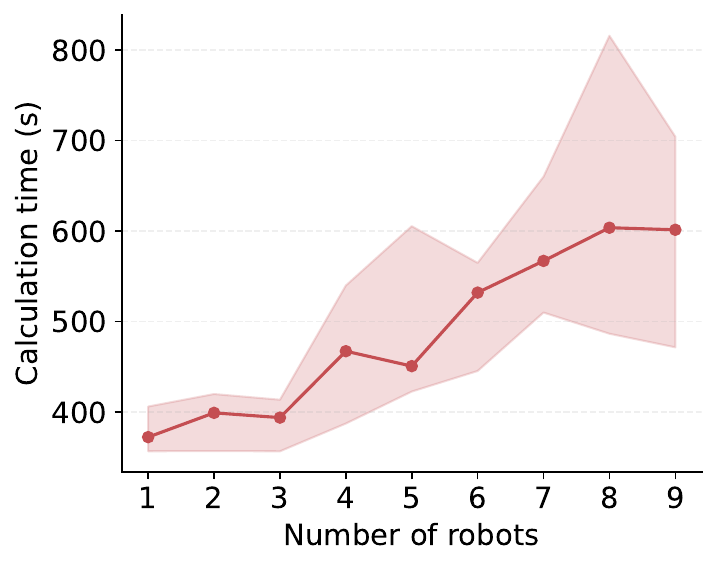}
        \caption{Makespan and Computation Time for Crate}
        \label{fig:scaling-crate}
    \end{subfigure}
    \hfill
    \begin{subfigure}[b]{\scalingwidth}
        \centering
        \includegraphics[height=\scalingheight,width=0.49\textwidth,keepaspectratio]{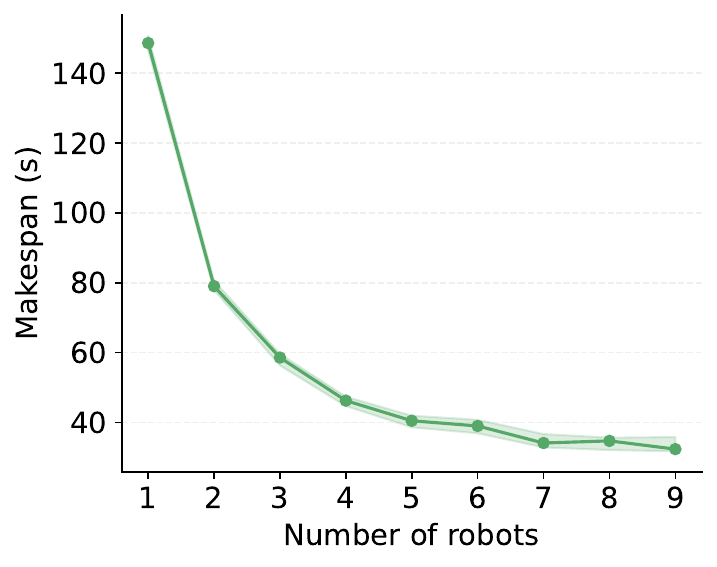}
        \includegraphics[height=\scalingheight,width=0.49\textwidth,keepaspectratio]{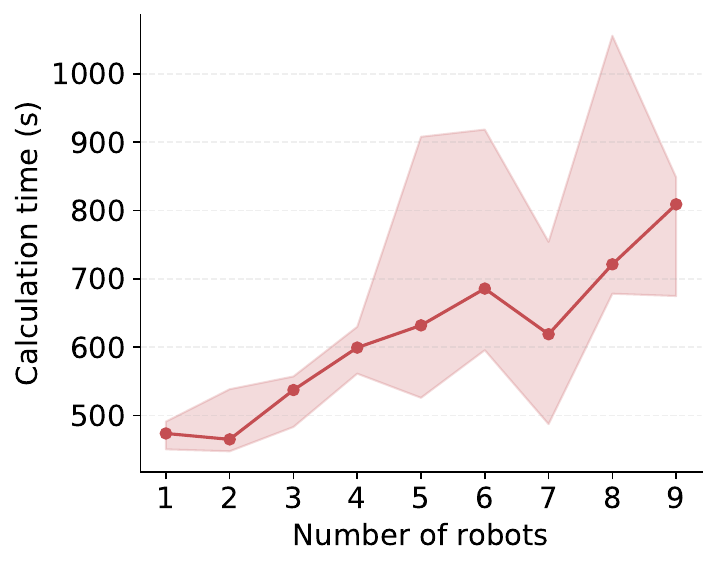}
        \caption{Makespan and Computation Time for Motor}
        \label{fig:scaling-motor}
    \end{subfigure}
    
    \vspace{0.8em}
    
    \begin{subfigure}[b]{\scalingwidth}
        \centering
        \includegraphics[height=\scalingheight,width=0.49\textwidth,keepaspectratio]{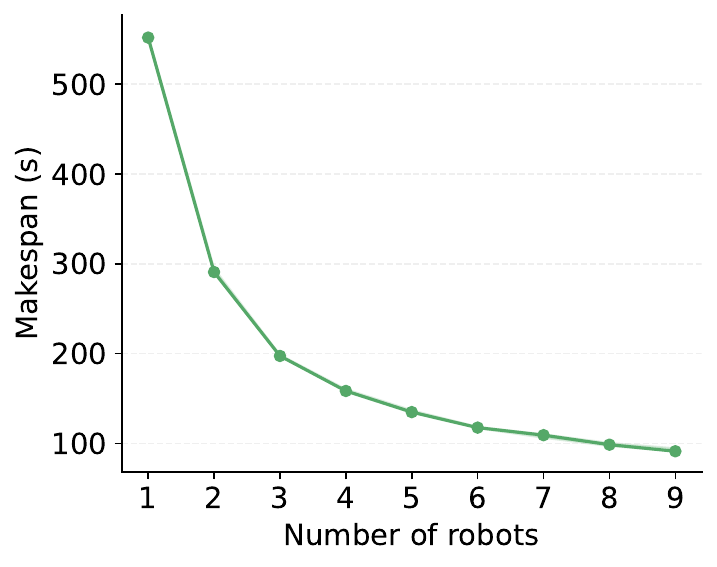}
        \includegraphics[height=\scalingheight,width=0.49\textwidth,keepaspectratio]{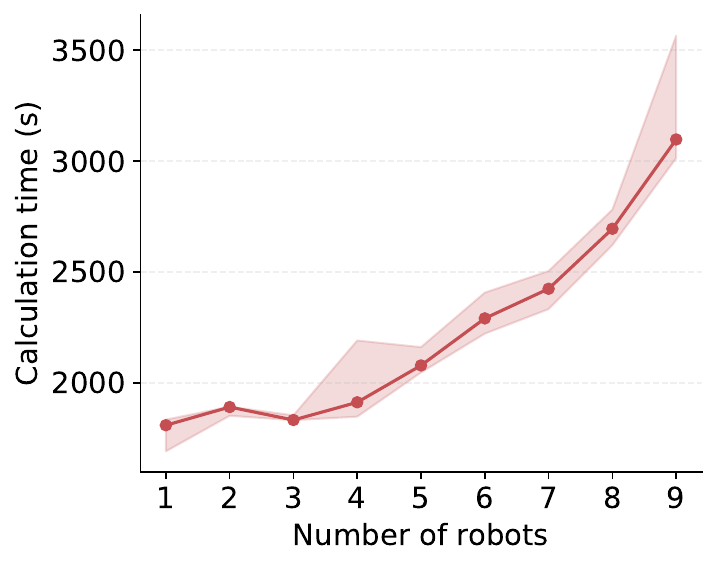}
        \caption{Makespan and Computation Time for Battery}
        \label{fig:scaling-battery}
    \end{subfigure}
    \hfill
    \begin{subfigure}[b]{\scalingwidth}
        \centering
        \includegraphics[height=\scalingheight,width=0.49\textwidth,keepaspectratio]{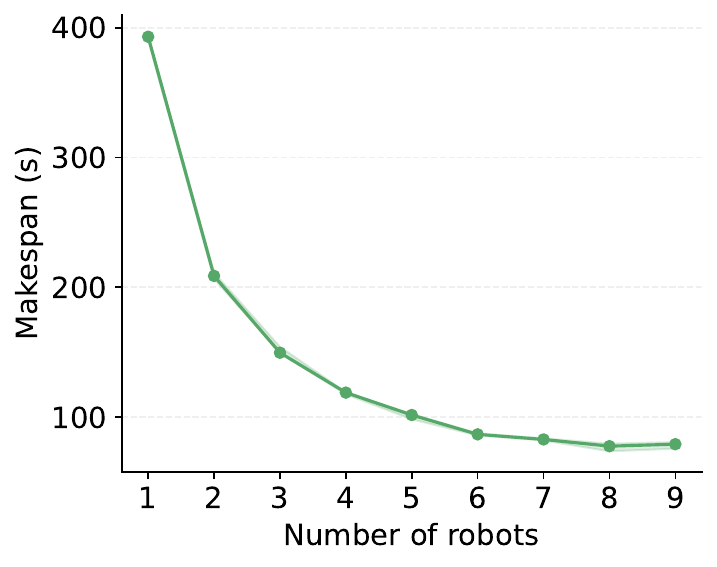}
        \includegraphics[height=\scalingheight,width=0.49\textwidth,keepaspectratio]{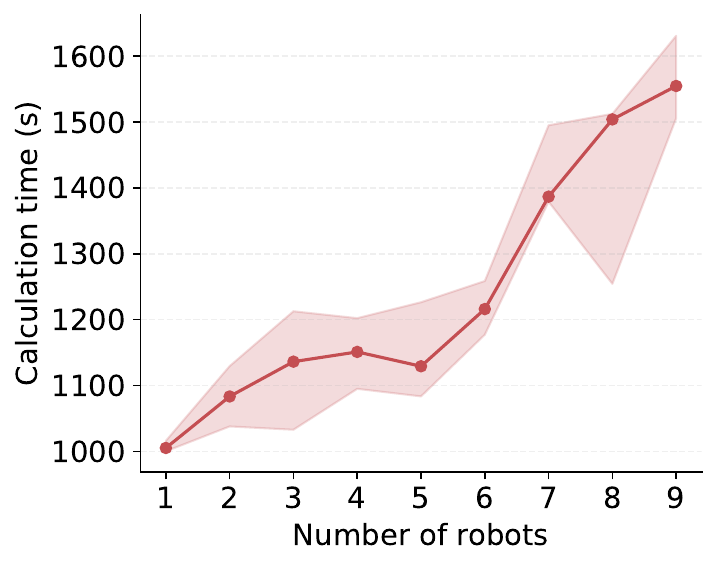}
        \caption{Makespan and Computation Time for Gearbox}
        \label{fig:scaling-gearbox}
    \end{subfigure}
    
    \caption{Scaling behavior for the two metrics of makespan and computation time for different robot counts in each disassembly
    scenario.\label{fig:calculation-time-and-makespan}}
\end{figure*}

%% file: src/figures/results_pareto_front.tex
\newcommand{\makespanheight}{4.5cm}     
\newcommand{\makespanwidth}{0.32\textwidth}   

\begin{figure*}[t]
    \centering
    
    \begin{subfigure}[b]{\makespanwidth}
        \centering
        \includegraphics[height=\makespanheight,width=\textwidth,keepaspectratio]{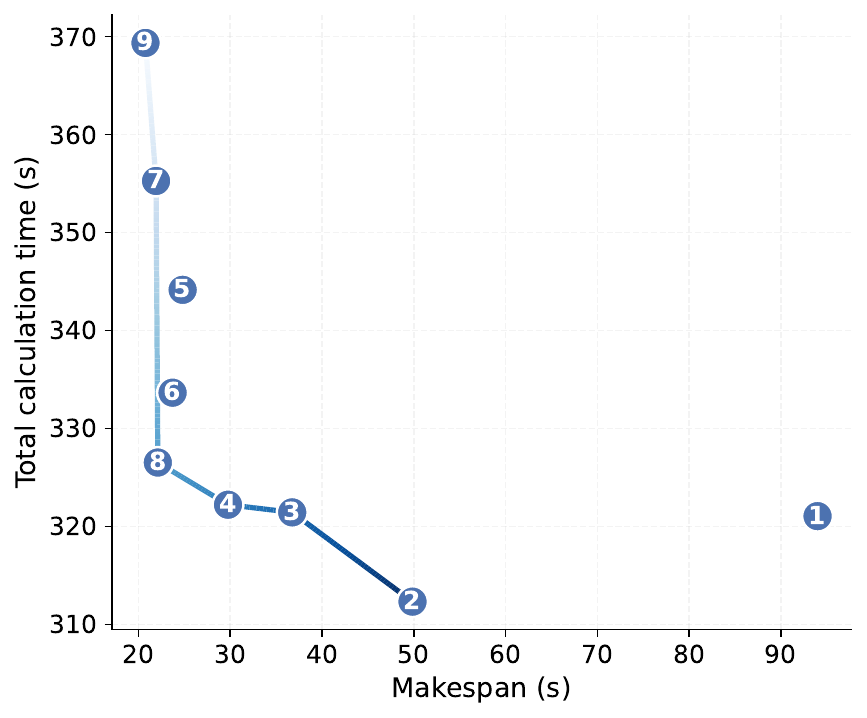}
        \caption{Pareto Front for Tower}
        \label{fig:makespan-tower}
    \end{subfigure}
    \hfill
    \begin{subfigure}[b]{\makespanwidth}
        \centering
        \includegraphics[height=\makespanheight,width=\textwidth,keepaspectratio]{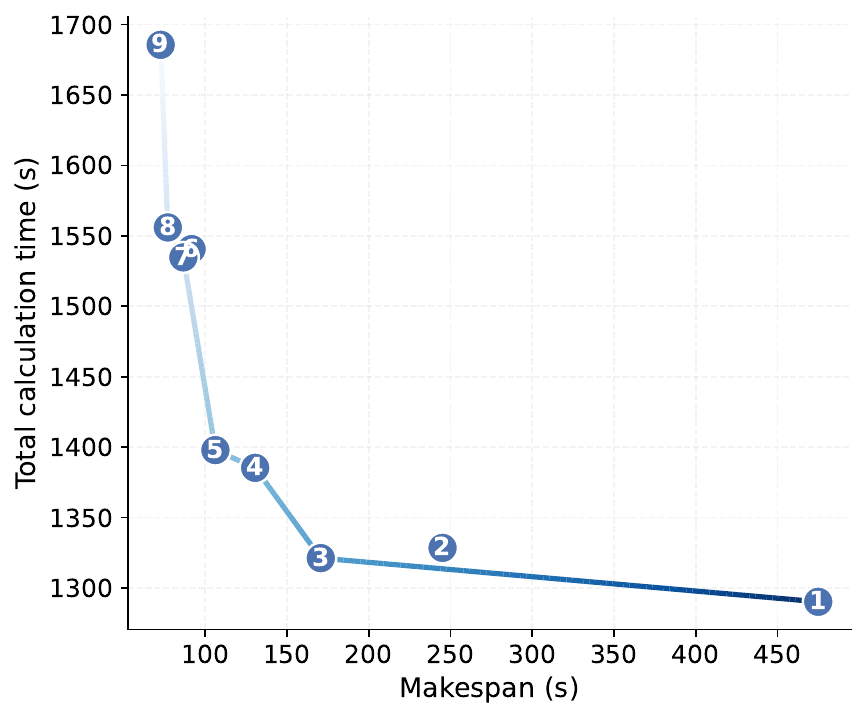}
        \caption{Pareto Front for Wall}
        \label{fig:makespan-wall}
    \end{subfigure}
    \hfill
    \begin{subfigure}[b]{\makespanwidth}
        \centering
        \includegraphics[height=\makespanheight,width=\textwidth,keepaspectratio]{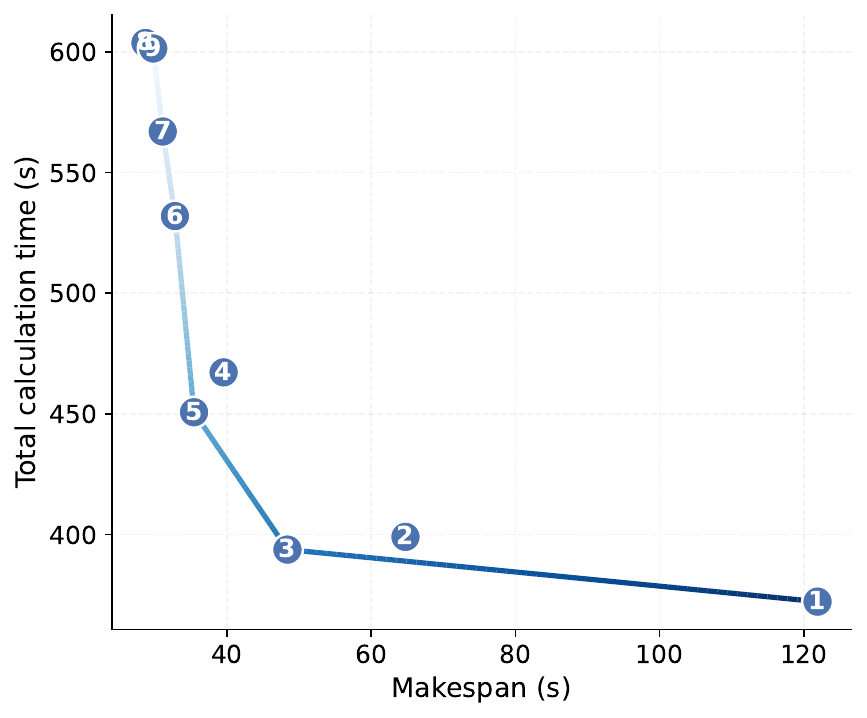}
        \caption{Pareto Front for Crate}
        \label{fig:makespan-crate}
    \end{subfigure}
    
    \vspace{0.8em}
    
    \begin{subfigure}[b]{\makespanwidth}
        \centering
        \includegraphics[height=\makespanheight,width=\textwidth,keepaspectratio]{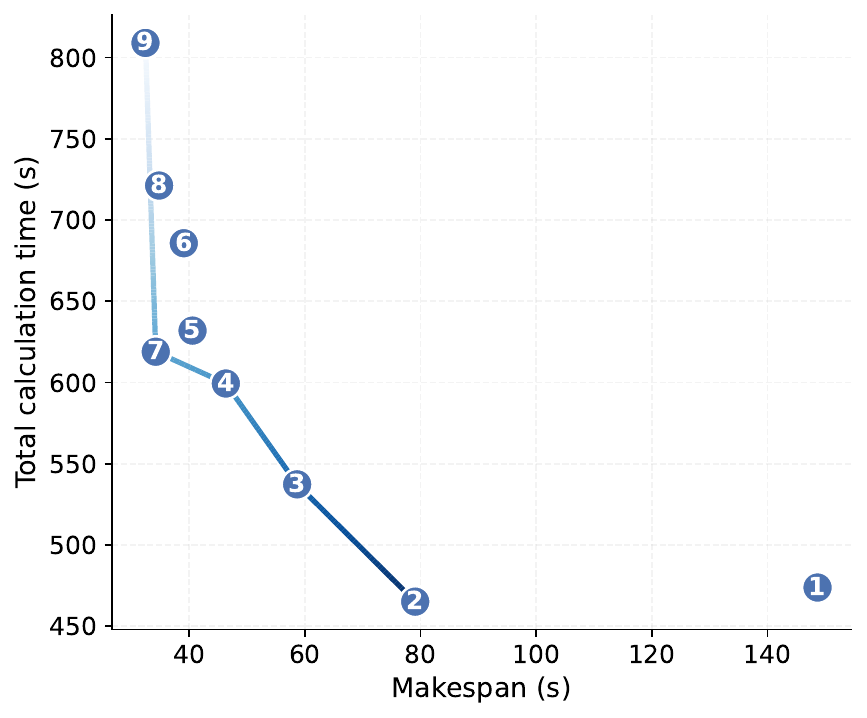}
        \caption{Pareto Front for Motor}
        \label{fig:makespan-motor}
    \end{subfigure}
    \hfill
    \begin{subfigure}[b]{\makespanwidth}
        \centering
        \includegraphics[height=\makespanheight,width=\textwidth,keepaspectratio]{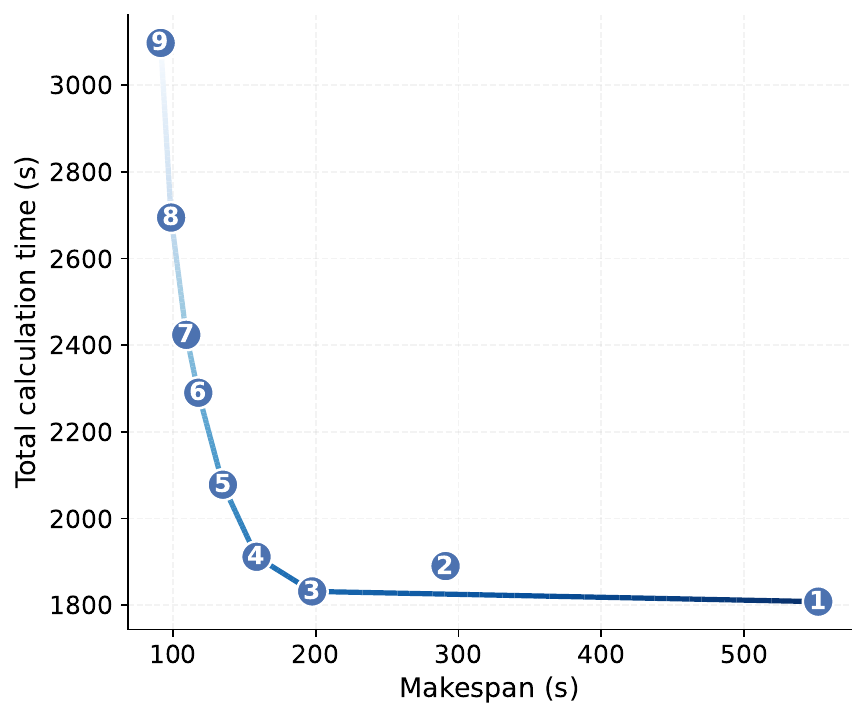}
        \caption{Pareto Front for Battery}
        \label{fig:makespan-battery}
    \end{subfigure}
    \hfill
    \begin{subfigure}[b]{\makespanwidth}
        \centering
        \includegraphics[height=\makespanheight,width=\textwidth,keepaspectratio]{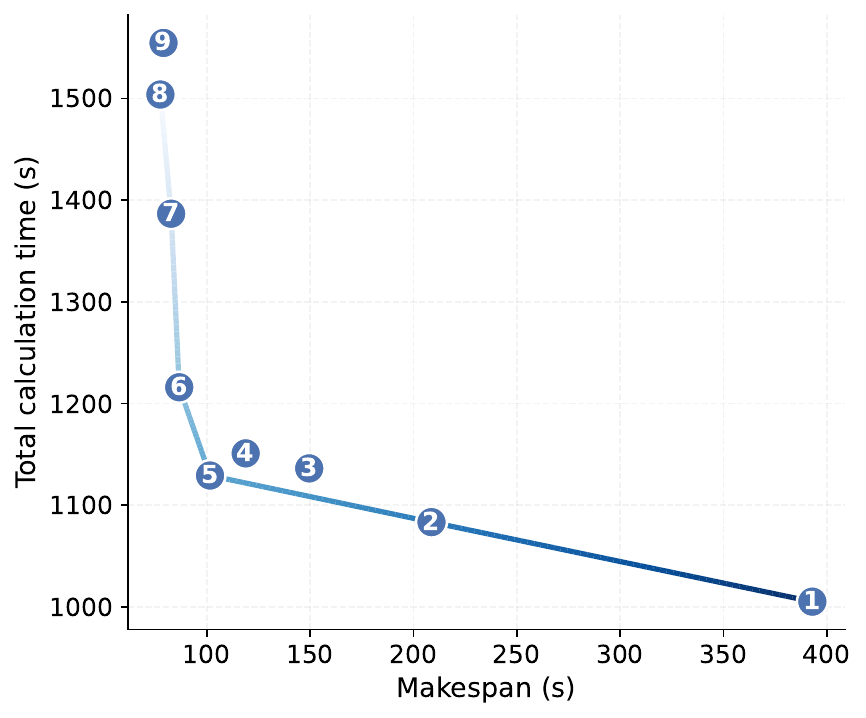}
        \caption{Pareto Front for Gearbox}
        \label{fig:makespan-gearbox}
    \end{subfigure}
    
    \caption{Pareto front for each disassembly
    scenario.\label{fig:pareto-front}}
\end{figure*}

%% file: src/figures/results_scheduling.tex
\newcommand{\timelineheight}{4.2cm}     
\newcommand{\timelinewidth}{0.49\textwidth}   

\begin{figure*}[!htbp]
    \centering
    
    \begin{subfigure}[b]{\timelinewidth}
        \centering
        \includegraphics[height=\timelineheight,width=\textwidth,keepaspectratio]{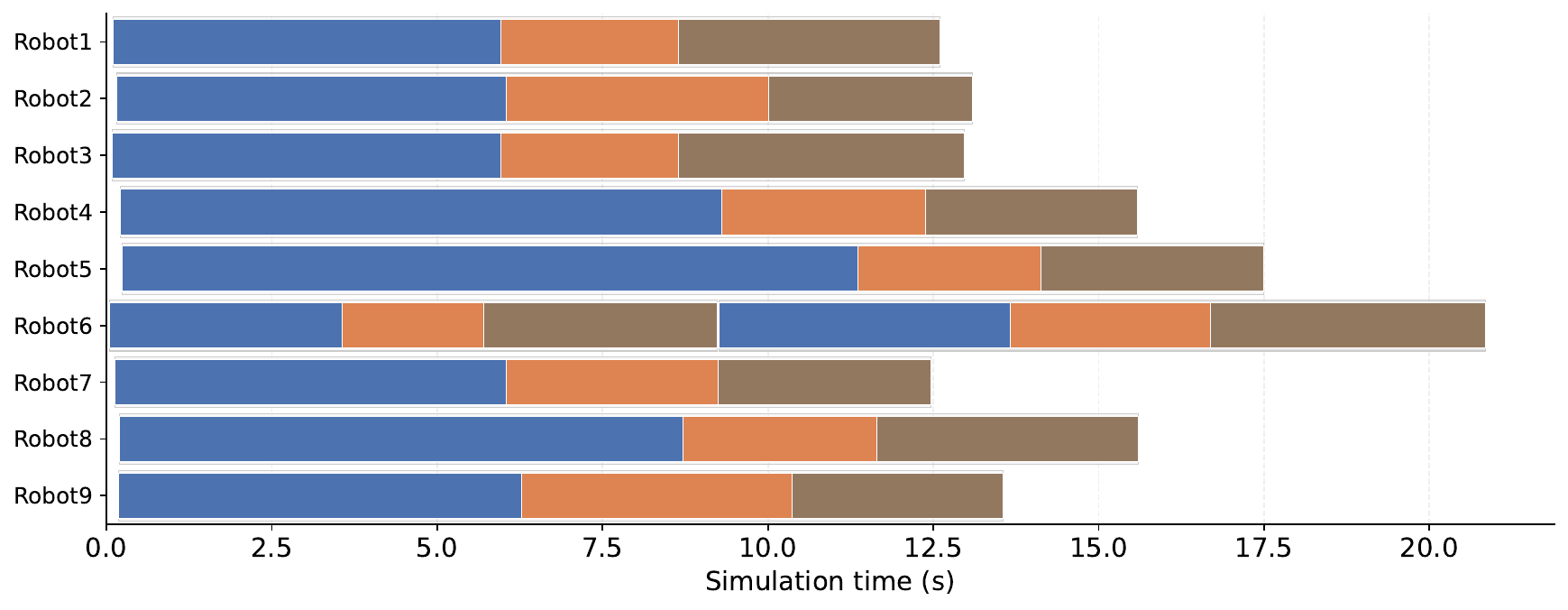}
        \caption{Tower}
        \label{fig:timeline-tower}
    \end{subfigure}
    \begin{subfigure}[b]{\timelinewidth}
        \centering
        \includegraphics[height=\timelineheight,width=\textwidth,keepaspectratio]{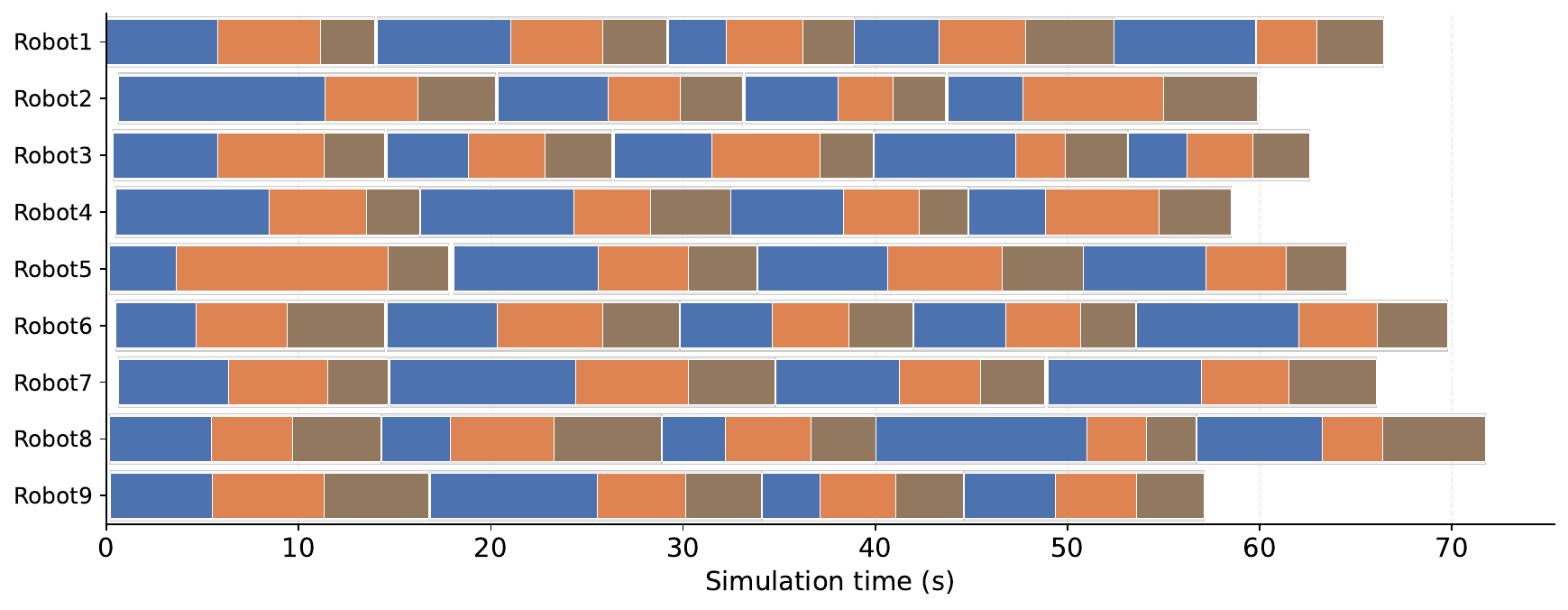}
        \caption{Wall}
        \label{fig:timeline-wall}
    \end{subfigure}    
    \begin{subfigure}[b]{\timelinewidth}
        \centering
        \includegraphics[height=\timelineheight,width=\textwidth,keepaspectratio]{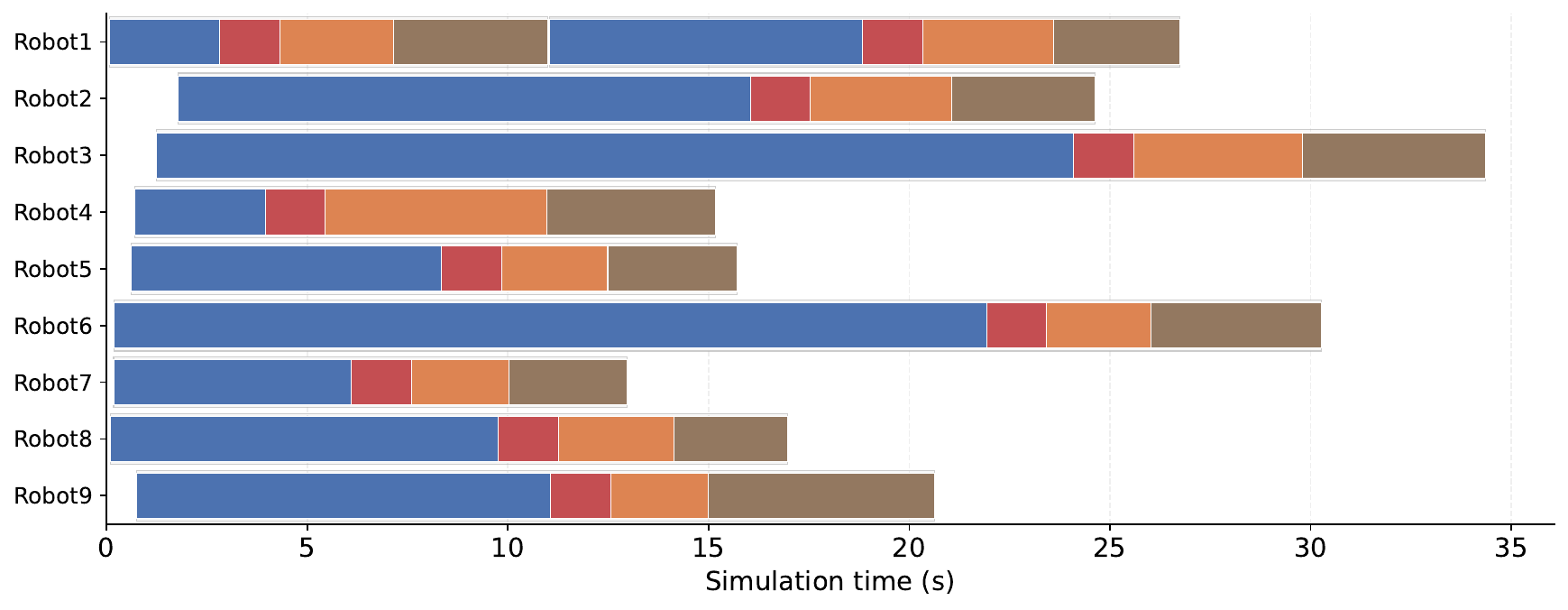}
        \caption{Crate}
        \label{fig:timeline-crate}
    \end{subfigure}
    \hfill
    \begin{subfigure}[b]{\timelinewidth}
        \centering
        \includegraphics[height=\timelineheight,width=\textwidth,keepaspectratio]{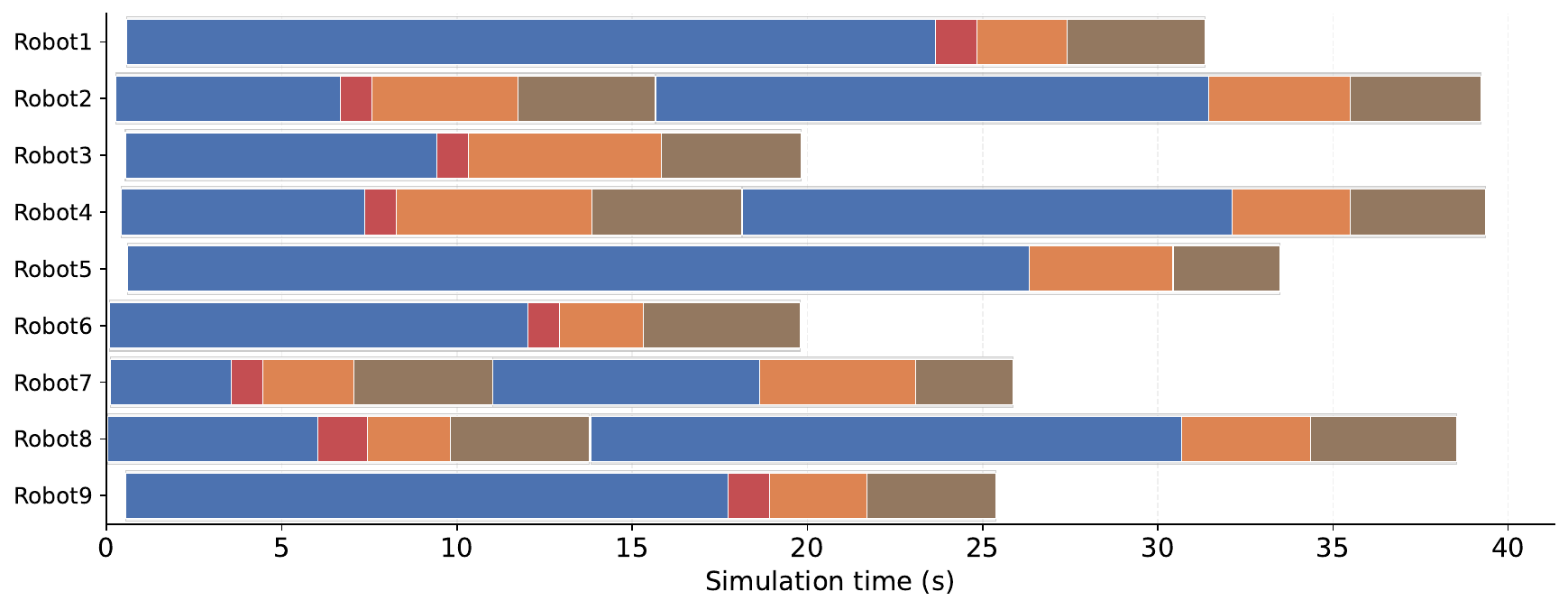}
        \caption{Motor}
        \label{fig:timeline-motor}
    \end{subfigure}
    
    \vspace{1.2em}
    
    \begin{subfigure}[b]{\timelinewidth}
        \centering
        \includegraphics[height=\timelineheight,width=\textwidth,keepaspectratio]{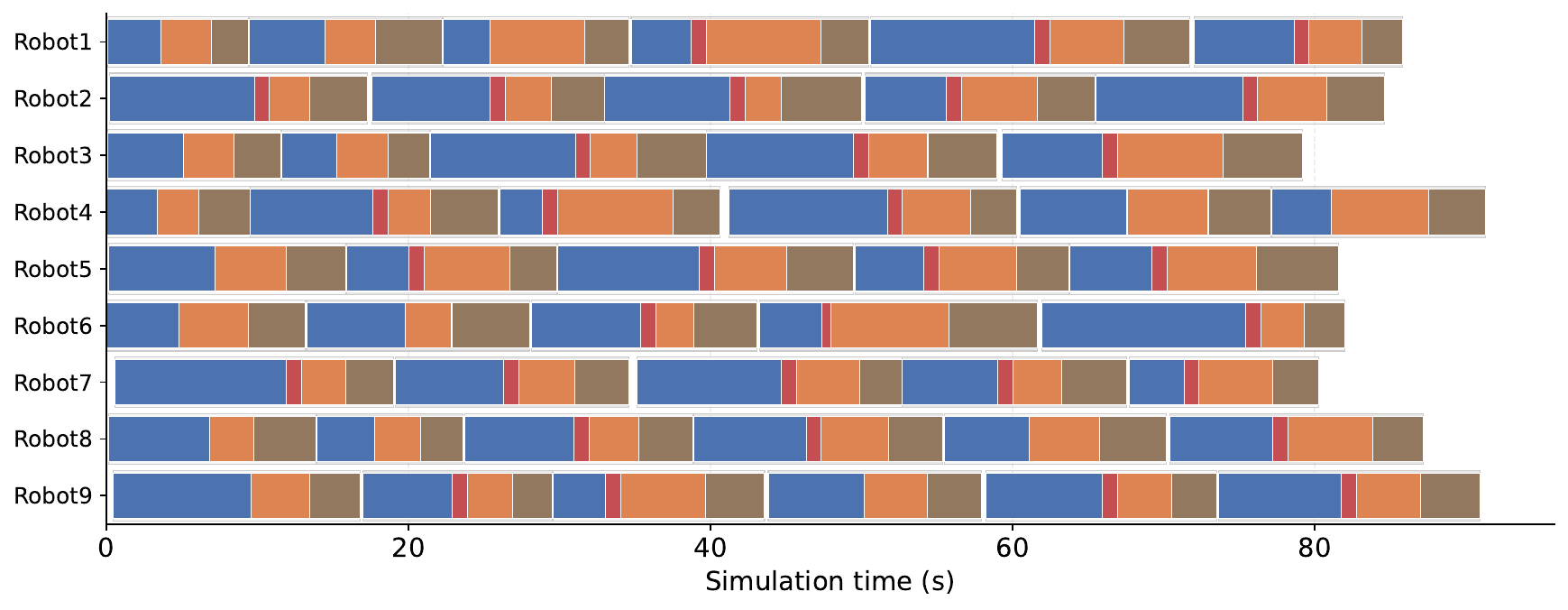}
        \caption{Battery}
        \label{fig:timeline-battery}
    \end{subfigure}
    \hfill
    \begin{subfigure}[b]{\timelinewidth}
        \centering
        \includegraphics[height=\timelineheight,width=\textwidth,keepaspectratio]{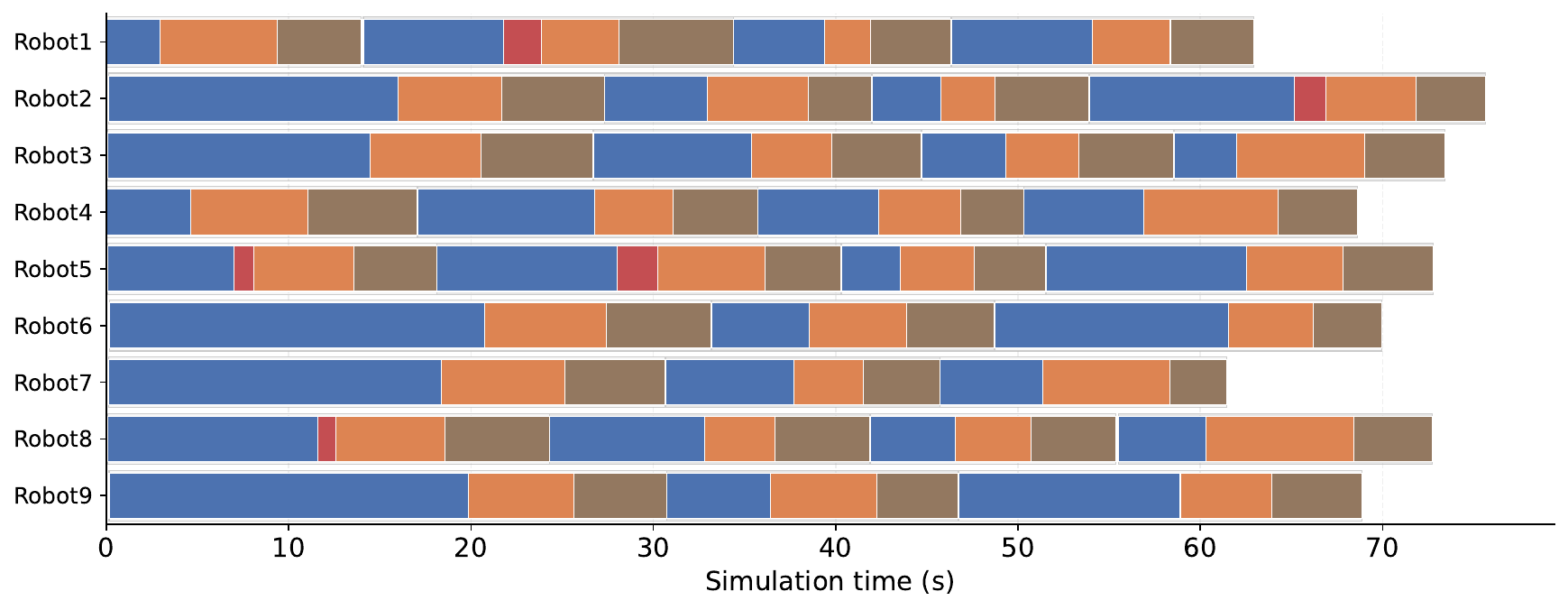}
        \caption{Gearbox}
        \label{fig:timeline-gearbox}
    \end{subfigure}
    
    \definecolor{lightblue}{RGB}{76, 114, 176}
    \definecolor{lightorange}{RGB}{221, 132, 82}
    \definecolor{lightred}{RGB}{196, 78, 82}
    \definecolor{lightbrown}{RGB}{147, 120, 96}

    \caption{Scheduling timelines of the disassembly process for each scenario. Tasks are color coded as \sqbox{lightblue} Pick,  \sqbox{lightorange} Place, \sqbox{lightred} Pull, and \sqbox{lightbrown} Exit.
    \label{fig:scheduling-timeline}}
\end{figure*}

%% file: src/figures/results_failure_analysis.tex
\newcommand{\failurerateheight}{4.2cm}     
\newcommand{\failureratewidth}{0.32\textwidth}   

\begin{figure*}[!htbp]
    \centering
    
    \begin{subfigure}[b]{\failureratewidth}
        \centering
        \includegraphics[height=\failurerateheight,width=\textwidth,keepaspectratio]{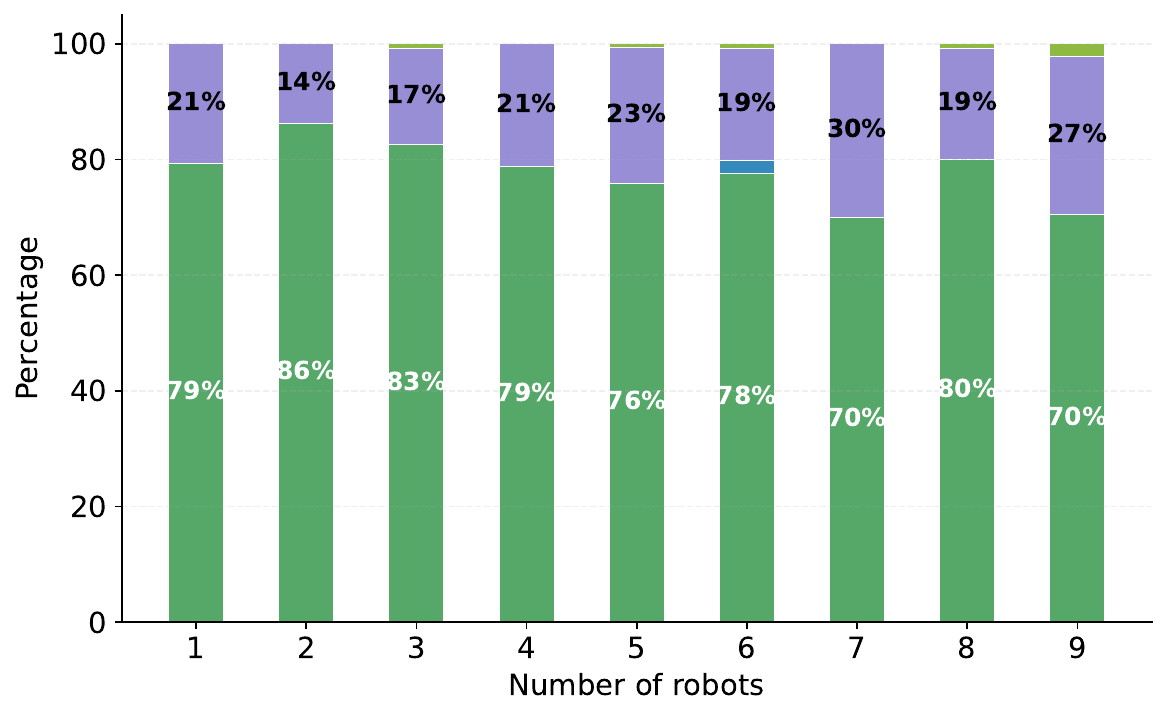}
        \caption{Tower}
        \label{fig:failure-tower}
    \end{subfigure}
    \begin{subfigure}[b]{\failureratewidth}
        \centering
        \includegraphics[height=\failurerateheight,width=\textwidth,keepaspectratio]{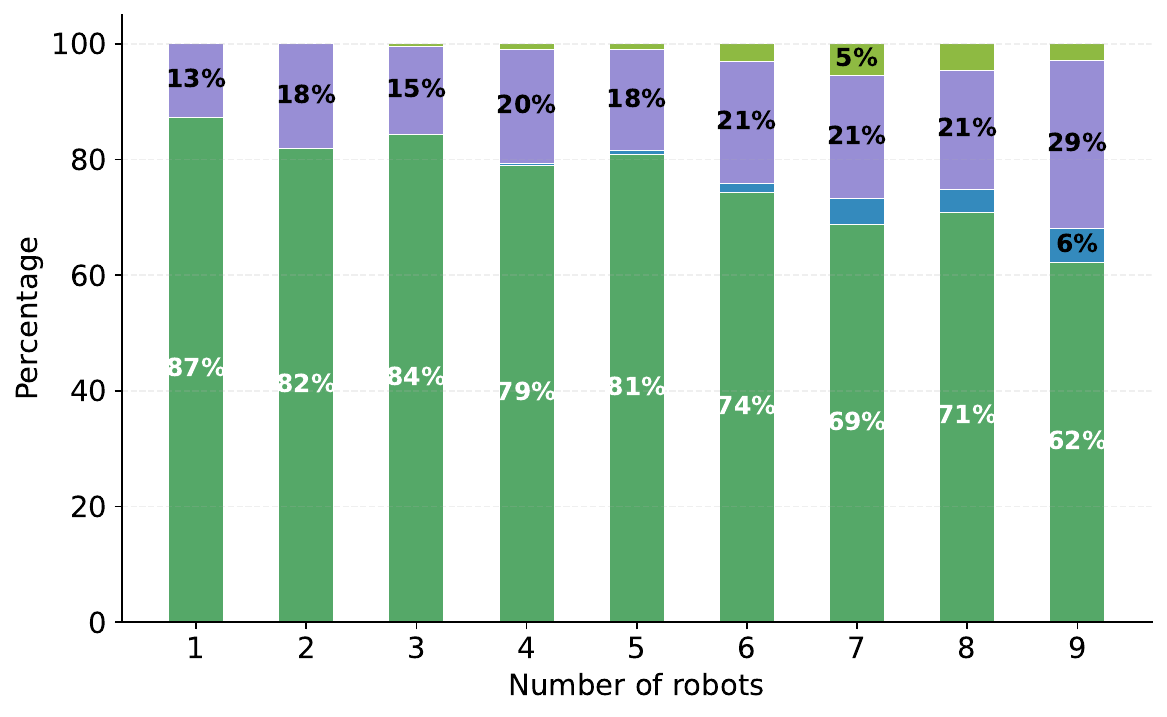}
        \caption{Wall}
        \label{fig:failure-wall}
    \end{subfigure}
    \begin{subfigure}[b]{\failureratewidth}
        \centering
        \includegraphics[height=\failurerateheight,width=\textwidth,keepaspectratio]{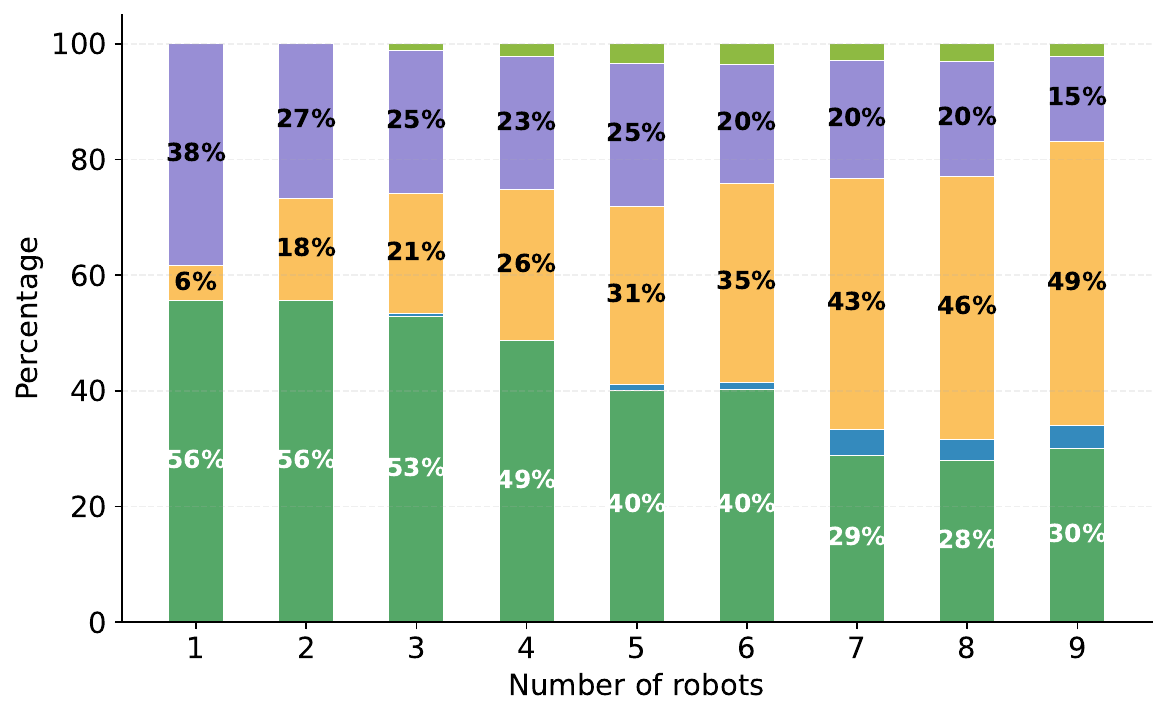}
        \caption{Crate}
        \label{fig:failure-crate}
    \end{subfigure}
    
    \begin{subfigure}[b]{\failureratewidth}
        \centering
        \includegraphics[height=\failurerateheight,width=\textwidth,keepaspectratio]{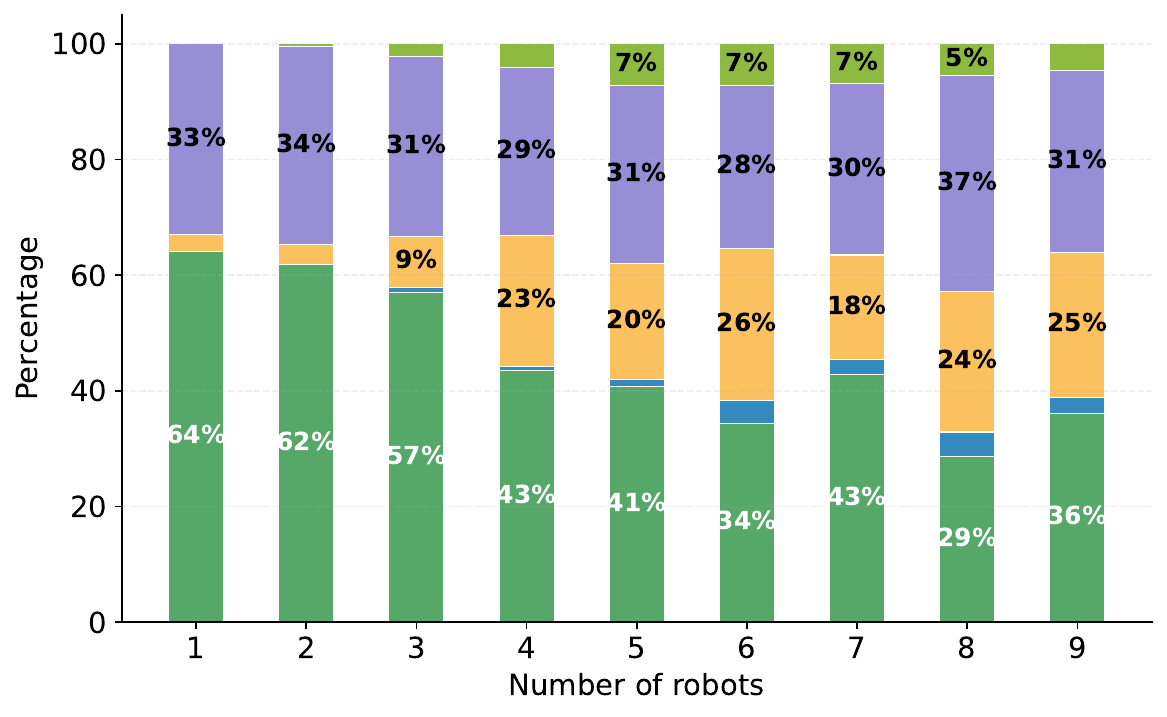}
        \caption{Motor}
        \label{fig:failure-motor}
    \end{subfigure}
    \begin{subfigure}[b]{\failureratewidth}
        \centering
        \includegraphics[height=\failurerateheight,width=\textwidth,keepaspectratio]{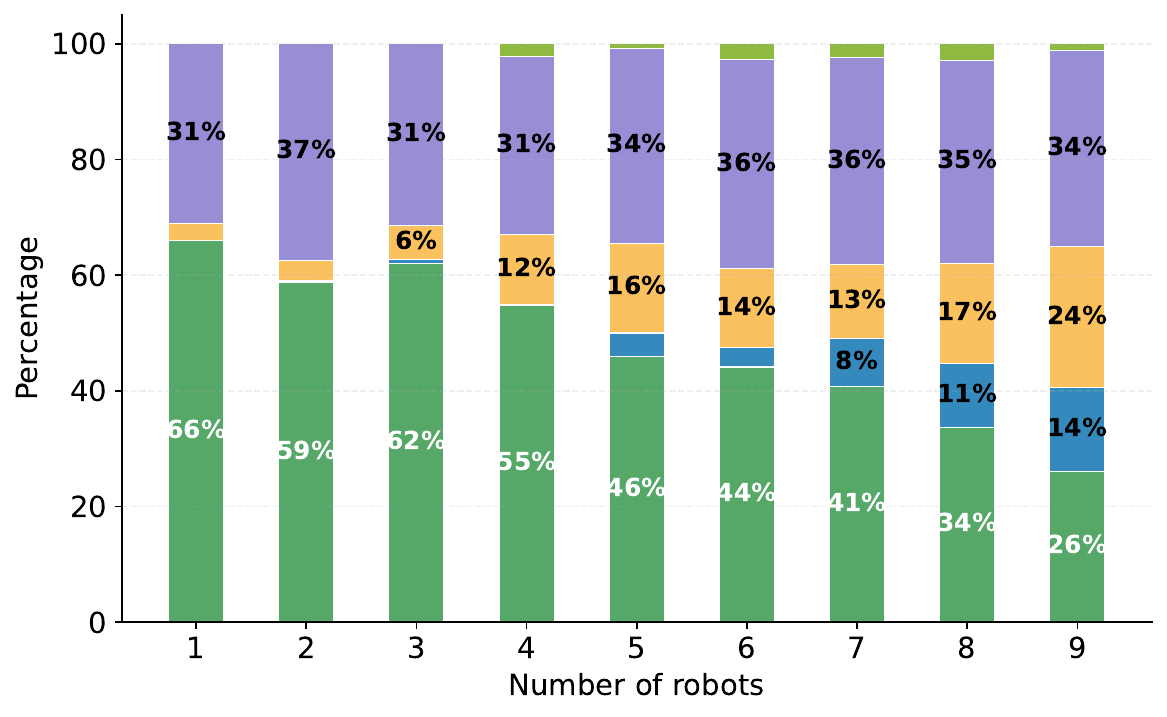}
        \caption{Battery}
        \label{fig:failure-battery}
    \end{subfigure}
    \begin{subfigure}[b]{\failureratewidth}
        \centering
        \includegraphics[height=\failurerateheight,width=\textwidth,keepaspectratio]{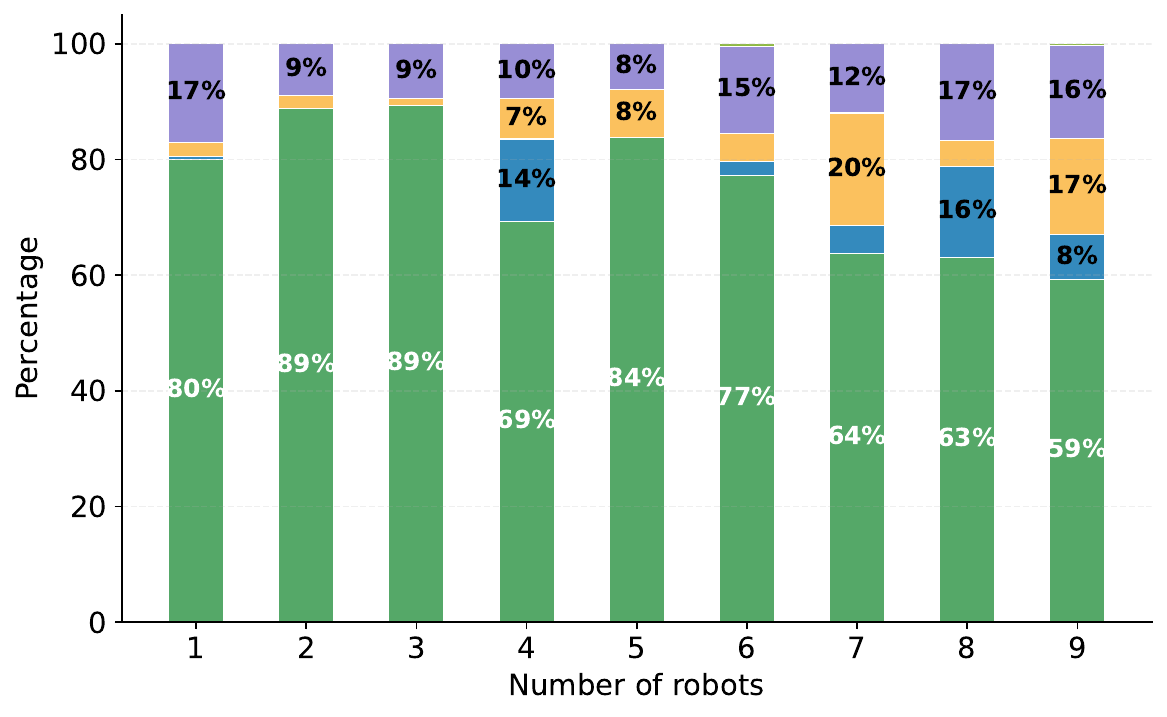}
        \caption{Gearbox}
        \label{fig:failure-gearbox}
    \end{subfigure}

    \definecolor{mypurple}{RGB}{152,142,213}
    \definecolor{myblue}{RGB}{52,138,189}
    \definecolor{myred}{RGB}{226,74,51}
    \definecolor{myyellow}{RGB}{251,193,94}
    \definecolor{mydarkgreen}{RGB}{85,168,104}
    \definecolor{mylightgreen}{RGB}{142,186,66}

    \caption{Failure rates across the different disassembly scenarios. \sqbox{mydarkgreen} Success \sqbox{mylightgreen} Exit fail \sqbox{myyellow} Pull fail \sqbox{myblue}   Plan to object fail  \sqbox{mypurple} Plan to goal fail
    \label{fig:results-failure-rates}
    }
    
\end{figure*}

%% file: src/figures/results_rrtstar_comparison.tex
\definecolor{compgreen}{RGB}{85,168,104}
\definecolor{compred}{RGB}{196,78,82}
\definecolor{compblue}{RGB}{76,114,176}
\begin{figure}[t]
  \centering
  \includegraphics[width=\linewidth, height=0.5\linewidth]{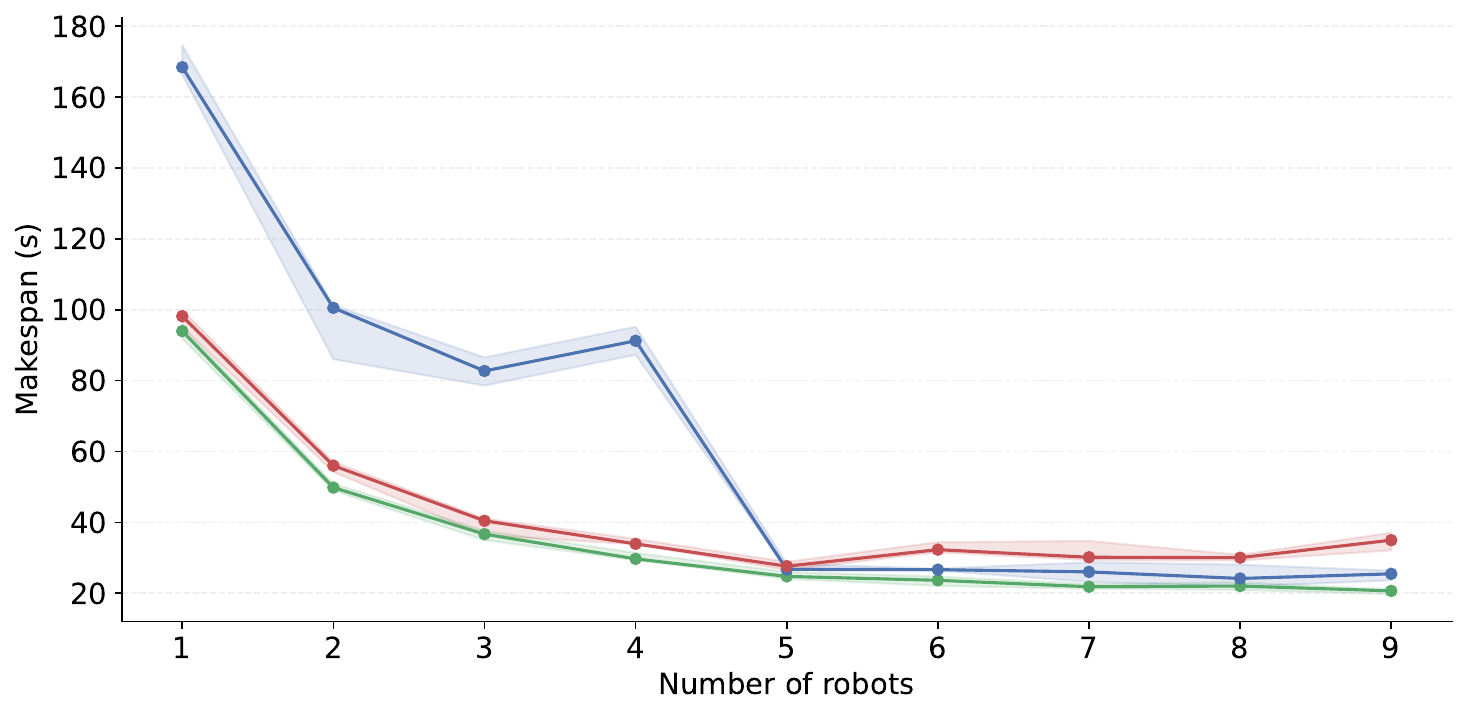}
  \caption{Makespan comparison between ST-RRT* and RRT*, showing \sqbox{compgreen} ST-RRT* , \sqbox{compblue} RRT* 5s, and \sqbox{compred} RRT* 10s.}
  \label{fig:rrtstar-compare-scaling}
\end{figure}

\begin{figure}[t]
  \centering
  \includegraphics[width=\linewidth]{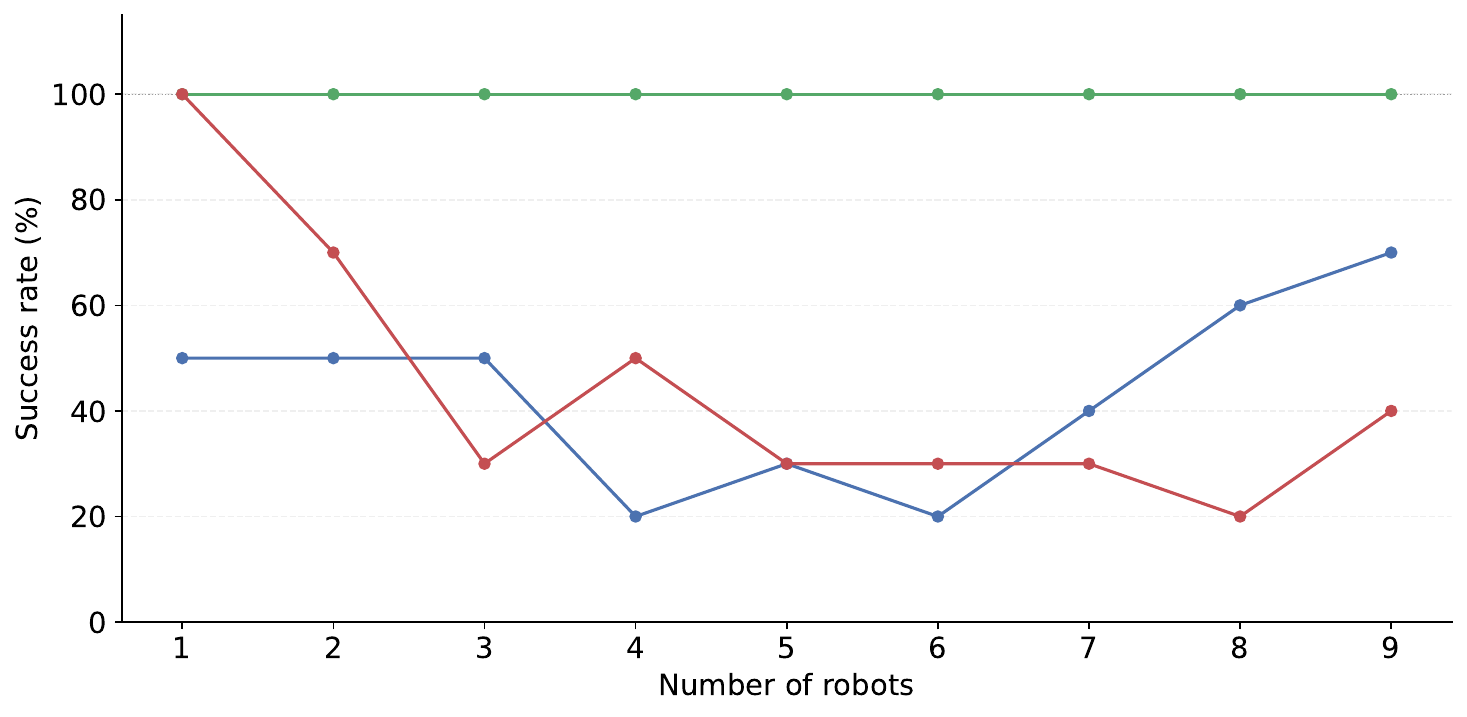}
  \caption{Success rate comparison of ST-RRT* and RRT*, showing \sqbox{compgreen} ST-RRT*, \sqbox{compblue} RRT* 5s, and \sqbox{compred} RRT* 10s.}
  \label{fig:rrtstar-compare-success-rate}
\end{figure}

%% file: src/07_conclusion.tex
\section{Discussion}

The experiments show two important results. 
First, CoMuDi can successfully coordinate multi-robot teams to disassemble large-scale objects with up to 49 pieces. 
Overall, we were able to show that CoMuDi can lower the makespan with an increase in the number of robots. 
This is a clear indicator that CoMuDi can efficiently orchestrate robot teams even when the number of robots increases, thereby decreasing the movement space for individual robots. 
While the computation time increases to coordinate larger robot teams, we have been able to show that almost all computations are close to the Pareto front, meaning we use the available computation time efficiently. 
Additionally, our results show that planning is a trade-off between makespan and computation time, thereby allowing users of the system to carefully weigh the number of robots needed to accomplish a certain disassembly task.

The second result concerns the comparison between ST-RRT* and RRT*. This comparison shows the importance of designing a planner to operate directly in space-time, rather than adapting an existing planner to incorporate this feature.
While RRT* with a ten second window achieves a comparable makespan to ST-RRT* for one to five robots, RRT* struggles to find competitive solutions beyond this.
The wider time window provides flexibility for detours, but the resulting state space becomes too large to explore and optimize once the workspace becomes crowded with other robots. 
The decline in success rate with an increasing number of robots reflects the same limitation.
ST-RRT* avoids this issue by allowing the time bound to shrink or grow on demand. It combines the makespan quality of the wider window at low robot counts with the sampling density of the narrower window at high robot counts, outperforming both across the full range. 
The adaptive bound comes at a modest computational cost, since additional work is required to determine the bounds at runtime.
This result shows that CoMuDi with ST-RRT* is a particularly efficient combination and leads consistently to successful results with a low makespan.

\subsection{Limitations}

While the results show that CoMuDi is approaching near optimal results in coordinating multiple robots, there are still some improvements possible to our framework. In particular:

\begin{itemize}
    \item \textbf{Depth-Aware Assignment Policy} : Currently, CoMuDi assigns the next
available object to a robot without considering the nodes currently executing in
the dependency graph. Taking this into account would balance the assembly more evenly throughout
the workspace. In the motor and gearbox scenario, a better assignment policy
could enable disassembly from two sides rather than relying on pure chance. This would prevent dependency chains with long pick phases, which slow down the
makespan. To tackle this, we need to consider the depth in the dependency graph,
the movement of other robots around the object, and take the available workspace into account.
\item \textbf{Clearance Cost} : The object extraction task considers each pull as a constrained motion scaled in time. This pull phase does not consider the proximity of
obstacles or the configurations of other robots planned earlier or later. A more
sophisticated approach would include a clearance cost to avoid critical areas so that the robot would proactively avoid obstructions for
future tasks while solving for the current pull. 
\item \textbf{Grouping of Objects} :
Another improvement concerns objects that are in close proximity. 
For example, the screws of the motor or the gears of the gearbox. Such objects could be grouped so that the IK solver can try to fit as many manipulators as possible into the workspace simultaneously.
    \item \textbf{Global Optimality} :
Another bottleneck is the division into separate planning problems. Since there is no overarching
planning objective, this can result in an unfavorable starting position in subsequent planning problems. The planner should be able to detect this and sacrifice local optimality
for global optimality.
\end{itemize}


\section{Conclusion}

We presented the coordinated multi-robot disassembly (CoMuDi) task and motion planning framework for disassembly tasks. CoMuDi integrates scale-invariant sampling~\cite{Bayraktar2026WAFR} for object extraction together with time-optimal motion planning~\cite{STRRTstar} into a coherent framework for multi-robot coordination~\cite{LongHorizon}. To ensure CoMuDi can robustly and efficiently solve disassembly tasks, we define generalized disassembly tasks, propagate temporal constraints, and develop a temporal collision checking framework. 

In six disassembly scenarios, we were able to show that CoMuDi can reliably solve disassembly tasks, even when those tasks require long-horizon planning with up to 9 robots and up to 49 objects in a single assembly. The experiments show that CoMuDi can successfully optimize makespan, especially when the number of robots is increased. The scheduling timelines show that robots minimize their idle time, and the ST-RRT* comparisons with RRT* show that CoMuDi with ST-RRT* achieves the best success rate and lowest makespan. 

While some limitations remain, the proposed planner already shows near optimal behavior and can be used to coordinate the motion of multi-robot teams operating in tight proximity to each other. This makes CoMuDi an essential part for any general-purpose disassembly system to tackle future applications in recycling and remanufacturing.

%% file: bib/general.bib
@article{LongHorizon,
  author    = {Hartmann, Valentin N. and Orthey, Andreas and Driess, Danny and Oguz, Ozgur S. and Toussaint, Marc},
  journal   = {IEEE Transactions on Robotics},
  title     = {Long-Horizon Multi-Robot Rearrangement Planning for Construction Assembly},
  year      = {2023},
  volume    = {39},
  number    = {1},
  pages     = {239--252},
}

@inproceedings{STRRTstar,
  author    = {Grothe, Francesco and Hartmann, Valentin N. and Orthey, Andreas and Toussaint, Marc},
  booktitle = {2022 International Conference on Robotics and Automation (ICRA)},
  title     = {{ST-RRT*}: Asymptotically-Optimal Bidirectional Motion Planning through Space-Time},
  year      = {2022},
  pages     = {3314--3320},
}

@article{dRRTjournal,
  author    = {Solovey, Kiril and Salzman, Oren and Halperin, Dan},
  title     = {Finding a Needle in an Exponential Haystack: Discrete {RRT} for Exploration of Implicit Roadmaps in Multi-Robot Motion Planning},
  journal   = {The International Journal of Robotics Research},
  volume    = {35},
  number    = {5},
  pages     = {501--513},
  year      = {2016},
}

@article{dRRTstar,
  author    = {Shome, Rahul and Solovey, Kiril and Dobson, Andrew and Halperin, Dan and Bekris, Kostas E.},
  title     = {{dRRT*}: Scalable and Informed Asymptotically-Optimal Multi-Robot Motion Planning},
  journal   = {Autonomous Robots},
  volume    = {44},
  number    = {3--4},
  pages     = {443--467},
  publisher = {Springer},
  year      = {2020},
}

@inproceedings{lgp,
  author    = {Toussaint, Marc},
  title     = {Logic-Geometric Programming: An Optimization-Based Approach to Combined Task and Motion Planning},
  year      = {2015},
  isbn      = {9781577357384},
  publisher = {AAAI Press},
  booktitle = {Proceedings of the 24th International Conference on Artificial Intelligence},
  pages     = {1930--1936},
  location  = {Buenos Aires, Argentina},
  series    = {IJCAI'15},
}

@article{assembleThemAll,
  title     = {Assemble Them All: Physics-Based Planning for Generalizable Assembly by Disassembly},
  author    = {Tian, Yunsheng and Xu, Jie and Li, Yichen and Luo, Jieliang and Sueda, Shinjiro and Li, Hui and Willis, Karl D. D. and Matusik, Wojciech},
  journal   = {ACM Transactions on Graphics},
  volume    = {41},
  number    = {6},
  articleno = {278},
  numpages  = {15},
  year      = {2022},
  publisher = {ACM},
}

@article{erdmann,
  author    = {Erdmann, Michael and Lozano-P{\'e}rez, Tom{\'a}s},
  title     = {On Multiple Moving Objects},
  journal   = {Algorithmica},
  volume    = {2},
  number    = {1},
  pages     = {477--521},
  year      = {1987},
}

@article{hsu2002randomized,
  author    = {Hsu, David and Kindel, Robert and Latombe, Jean-Claude and Rock, Stephen},
  title     = {Randomized Kinodynamic Motion Planning with Moving Obstacles},
  journal   = {The International Journal of Robotics Research},
  volume    = {21},
  number    = {3},
  pages     = {233--255},
  year      = {2002},
}

@article{karaman2011sampling,
  author    = {Karaman, Sertac and Frazzoli, Emilio},
  title     = {Sampling-Based Algorithms for Optimal Motion Planning},
  journal   = {The International Journal of Robotics Research},
  volume    = {30},
  number    = {7},
  pages     = {846--894},
  year      = {2011},
}

@inproceedings{phillips2011sipp,
  author    = {Phillips, Mike and Likhachev, Maxim},
  title     = {{SIPP}: Safe Interval Path Planning for Dynamic Environments},
  booktitle = {Proceedings of the IEEE International Conference on Robotics and Automation (ICRA)},
  pages     = {5628--5635},
  year      = {2011},
}

@article{andreychuk2022multi,
  author    = {Andreychuk, Anton and Yakovlev, Konstantin and Surynek, Pavel and Atzmon, Dor and Stern, Roni},
  title     = {Multi-Agent Pathfinding with Continuous Time},
  journal   = {Artificial Intelligence},
  volume    = {305},
  pages     = {103662},
  year      = {2022},
}

@article{cap2015prioritized,
  author    = {{\v{C}}{\'a}p, Michal and Nov{\'a}k, Peter and Kleiner, Alexander and Seleck{\'y}, Martin},
  title     = {Prioritized Planning Algorithms for Trajectory Coordination of Multiple Mobile Robots},
  journal   = {IEEE Transactions on Automation Science and Engineering},
  volume    = {12},
  number    = {3},
  pages     = {835--849},
  year      = {2015},
}

@article{Sharon2015CBS,
title = {Conflict-based search for optimal multi-agent pathfinding},
journal = {Artificial Intelligence},
volume = {219},
pages = {40-66},
year = {2015},
author = {Guni Sharon and Roni Stern and Ariel Felner and Nathan R. Sturtevant},
}

@inproceedings{Kaelbling2011TAMP,
  title={Hierarchical task and motion planning in the now},
  author={Leslie Pack Kaelbling and Tomas Lozano-Perez},
  booktitle={2011 IEEE International Conference on Robotics and Automation},
  year={2011},
  pages={1470-1477},
}

@INPROCEEDINGS{scrivastava2014Combined,
  author={Srivastava, Siddharth and Fang, Eugene and Riano, Lorenzo and Chitnis, Rohan and Russell, Stuart and Abbeel, Pieter},
  booktitle={2014 IEEE International Conference on Robotics and Automation (ICRA)}, 
  title={Combined task and motion planning through an extensible planner-independent interface layer}, 
  year={2014},
  volume={},
  number={},
  pages={639-646},
}

@inproceedings{garrett2020pddlstream,
  title={Pddlstream: Integrating symbolic planners and blackbox samplers via optimistic adaptive planning},
  author={Garrett, Caelan Reed and Lozano-P{\'e}rez, Tom{\'a}s and Kaelbling, Leslie Pack},
  booktitle={Proceedings of the international conference on automated planning and scheduling},
  volume={30},
  pages={440--448},
  year={2020}
}

@techreport{PDDL,
author = {Ghallab, Malik and Knoblock, Craig and Wilkins, David and Barrett, Anthony and Christianson, Dave and Friedman, Marc and Kwok, Chung and Golden, Keith and Penberthy, Scott and Smith, David and Sun, Ying and Weld, Daniel},
institution = {Yale Center for Computational Vision and Control},
number = {CVC TR-98-003/DCS TR-1165},
year = {1998},
month = {08},
title = {PDDL - The Planning Domain Definition Language}
}

@INPROCEEDINGS{toussaint2017MultiBound,
  author={Toussaint, Marc and Lopes, Manuel},
  booktitle={2017 IEEE International Conference on Robotics and Automation (ICRA)}, 
  title={Multi-bound tree search for logic-geometric programming in cooperative manipulation domains}, 
  year={2017},
  volume={},
  number={},
  pages={4044-4051},
}

@inproceedings{hartmann2023towards,
  author    = {Hartmann, Valentin N. and Toussaint, Marc},
  title     = {Towards Computing Low-Makespan Solutions for Multi-Arm Multi-Task Planning Problems},
  booktitle = {ICAPS Workshop on Planning and Robotics (PlanRob)},
  year      = {2023},
  eprint    = {2305.17527},
  archivePrefix = {arXiv},
  primaryClass  = {cs.RO}
}

@article{chen2022CoopTAMP,
  author        = {Chen, Jingkai and Li, Jiaoyang and Huang, Yijiang and Garrett, Caelan and Sun, Dawei and Fan, Chuchu and Hofmann, Andreas and Mueller, Caitlin and Koenig, Sven and Williams, Brian C.},
  title         = {Cooperative Task and Motion Planning for Multi-Arm Assembly Systems},
  journal       = {arXiv preprint arXiv:2203.02475},
  year          = {2022},
  eprint        = {2203.02475},
  archivePrefix = {arXiv},
  primaryClass  = {cs.RO}
}

@article{lambert2003DisassemblySequencing,
author = {Lambert, A.J.D.},
year = {2003},
month = {11},
pages = {3721-3759},
title = {Disassembly sequencing: A survey},
volume = {41},
journal = {International Journal of Production Research},
}

@article{poschmann2021fostering,
  title={Fostering end-of-life utilization by information-driven robotic disassembly},
  author={Poschmann, Hendrik and Br{\"u}ggemann, Holger and Goldmann, Daniel},
  journal={Procedia CIRP},
  volume={98},
  pages={282--287},
  year={2021},
  publisher={Elsevier}
}

@article{kheder2014SequenceGeneticAlgo,
author = {Kheder, Maroua and Trigui, M. and Aifaoui, Nizar},
year = {2014},
month = {11},
title = {Disassembly sequence planning based on a genetic algorithm},
volume = {229},
journal = {Proceedings of the Institution of Mechanical Engineers, Part C: Journal of Mechanical Engineering Science},
}

@inproceedings{sintov2014timebased,
  author    = {Sintov, Avishai and Shapiro, Amir},
  title     = {Time-Based {RRT} Algorithm for Rendezvous Planning of Two Dynamic Systems},
  booktitle = {Proceedings of the {IEEE} International Conference on Robotics and Automation ({ICRA})},
  pages     = {6745--6750},
  year      = {2014},
}

@article{vandenberg2005roadmap,
  author  = {van den Berg, J. and Overmars, Mark H.},
  title   = {Roadmap-Based Motion Planning in Dynamic Environments},
  journal = {IEEE Transactions on Robotics},
  volume  = {21},
  number  = {5},
  pages   = {885--897},
  year    = {2005},
}

@article{vandenberg2008timeoptimal,
  author  = {van den Berg, J. and Overmars, Mark H.},
  title   = {Planning Time-Minimal Safe Paths Amidst Unpredictably Moving Obstacles},
  journal = {The International Journal of Robotics Research},
  volume  = {27},
  number  = {11-12},
  pages   = {1274--1294},
  year    = {2008},
}

@inproceedings{ma2019pbs,
  author  = {Ma, Hang and Harabor, Daniel and Stuckey, Peter J. and Li, Jiaoyang and Koenig, Sven},
  title   = {Searching with Consistent Prioritization for Multi-Agent Path Finding},
  booktitle = {Proceedings of the {AAAI} Conference on Artificial Intelligence},
  volume  = {33},
  number  = {1},
  pages   = {7643--7650},
  year    = {2019},
}

@inproceedings{yu2013intractability,
  author  = {Yu, Jingjin and LaValle, Steven M.},
  title   = {Structure and Intractability of Optimal Multi-Robot Path Planning on Graphs},
  booktitle = {Proceedings of the {AAAI} Conference on Artificial Intelligence},
  volume  = {27},
  number  = {1},
  pages   = {1443--1449},
  year    = {2013},
}

@article{bennewitz2002finding,
  author  = {Bennewitz, Maren and Burgard, Wolfram and Thrun, Sebastian},
  title   = {Finding and Optimizing Solvable Priority Schemes for Decoupled Path Planning Techniques for Teams of Mobile Robots},
  journal = {Robotics and Autonomous Systems},
  volume  = {41},
  number  = {2-3},
  pages   = {89--99},
  year    = {2002},
}

@article{garrett2021survey,
  author  = {Garrett, Caelan Reed and Chitnis, Rohan and Holladay, Rachel and Kim, Beomjoon and Silver, Tom and Kaelbling, Leslie Pack and Lozano-P{\'e}rez, Tom{\'a}s},
  title   = {Integrated Task and Motion Planning},
  journal = {Annual Review of Control, Robotics, and Autonomous Systems},
  volume  = {4},
  pages   = {265--293},
  year    = {2021},
}

@article{simeon2004manipulation,
  author  = {Sim{\'e}on, Thierry and Laumond, Jean-Paul and Cort{\'e}s, Juan and Sahbani, Anis},
  title   = {Manipulation Planning with Probabilistic Roadmaps},
  journal = {The International Journal of Robotics Research},
  volume  = {23},
  number  = {7-8},
  pages   = {729--746},
  year    = {2004},
}

@article{hauser2010multimodal,
  author  = {Hauser, Kris and Latombe, Jean-Claude},
  title   = {Multi-Modal Motion Planning in Non-Expansive Spaces},
  journal = {The International Journal of Robotics Research},
  volume  = {29},
  number  = {7},
  pages   = {897--915},
  year    = {2010},
}

@inproceedings{krontiris2015rearrangement,
  author    = {Krontiris, Athanasios and Bekris, Kostas E.},
  title     = {Dealing with Difficult Instances of Object Rearrangement},
  booktitle = {Proceedings of Robotics: Science and Systems ({RSS})},
  year      = {2015},
}

@incollection{vegabrown2016asymptotically,
  author    = {Vega-Brown, William and Roy, Nicholas},
  title     = {Asymptotically Optimal Planning under Piecewise-Analytic Constraints},
  booktitle = {Algorithmic Foundations of Robotics {XII} ({WAFR} 2016)},
  series    = {Springer Proceedings in Advanced Robotics},
  volume    = {13},
  pages     = {528--543},
  publisher = {Springer},
  year      = {2020},
}

@incollection{toussaint2017komo,
  author    = {Toussaint, Marc},
  title     = {A Tutorial on {N}ewton Methods for Constrained Trajectory Optimization and Relations to {SLAM}, {G}aussian Process Smoothing, Optimal Control, and Probabilistic Inference},
  booktitle = {Geometric and Numerical Foundations of Movements},
  series    = {Springer Tracts in Advanced Robotics},
  volume    = {117},
  pages     = {361--392},
  publisher = {Springer},
  year      = {2017},
}

@inproceedings{toussaint2018differentiable,
  author    = {Toussaint, Marc and Allen, Kelsey R. and Smith, Kevin A. and Tenenbaum, Joshua B.},
  title     = {Differentiable Physics and Stable Modes for Tool-Use and Manipulation Planning},
  booktitle = {Proceedings of Robotics: Science and Systems ({RSS})},
  year      = {2018},
}

@article{sucan2012the-open-motion-planning-library,
    Author = {Ioan A. {\c{S}}ucan and Mark Moll and Lydia E. Kavraki},
    Journal = {{IEEE} Robotics \& Automation Magazine},
    Month = {December},
    Number = {4},
    Pages = {72--82},
    Title = {The {O}pen {M}otion {P}lanning {L}ibrary},
    Volume = {19},
    Year = {2012}
}

@INPROCEEDINGS{pan2021mrTAMPframework,
  author={Pan, Tianyang and Wells, Andrew M. and Shome, Rahul and Kavraki, Lydia E.},
  booktitle={2021 IEEE/RSJ International Conference on Intelligent Robots and Systems (IROS)}, 
  title={A General Task and Motion Planning Framework For Multiple Manipulators}, 
  year={2021},
  volume={},
  number={},
  pages={3168-3174},
}

@article{kongar2006ga,
  author  = {Kongar, Elif and Gupta, Surendra M.},
  title   = {Disassembly Sequencing Using Genetic Algorithm},
  journal = {The International Journal of Advanced Manufacturing Technology},
  volume  = {30},
  number  = {5-6},
  pages   = {497--506},
  year    = {2006},
}

@article{smith2012dssg,
  author  = {Smith, Shana and Smith, Gregory and Chen, Wei-Hua},
  title   = {Disassembly Sequence Structure Graphs: An Optimal Approach for Multiple-Target Selective Disassembly Sequence Planning},
  journal = {Advanced Engineering Informatics},
  volume  = {26},
  number  = {2},
  pages   = {306--316},
  year    = {2012},
}

@article{lazzerini2000ga,
  author  = {Lazzerini, Beatrice and Marcelloni, Francesco},
  title   = {A Genetic Algorithm for Generating Optimal Assembly Plans},
  journal = {Artificial Intelligence in Engineering},
  volume  = {14},
  number  = {4},
  pages   = {319--329},
  year    = {2000},
}

@INPROCEEDINGS{wagner2011mstar,
  author={Wagner, Glenn and Choset, Howie},
  booktitle={2011 IEEE/RSJ International Conference on Intelligent Robots and Systems}, 
  title={M*: A complete multirobot path planning algorithm with performance bounds}, 
  year={2011},
  volume={},
  number={},
  pages={3260-3267},
}

@inproceedings{Bayraktar2026WAFR,
	author = {Bayraktar, Servet B. and Orthey, Andreas and Toussaint, Marc},
	title = {Scale-Invariant Sampling in Multi-Arm Bandit Motion Planning for Object Extraction},
	booktitle = {World Symposium on the Algorithmic Foundations of Robotics},
	year = {2026},
}
